\documentclass{article} %
\usepackage{dex2hoi_preprint,times}

\usepackage{amsmath,amsfonts,bm}

\def\eqref#1{equation~\ref{#1}}

\def\1{\bm{1}}

\DeclareMathAlphabet{\mathsfit}{\encodingdefault}{\sfdefault}{m}{sl}
\SetMathAlphabet{\mathsfit}{bold}{\encodingdefault}{\sfdefault}{bx}{n}

\usepackage{hyperref}
\usepackage{url}
\usepackage[utf8]{inputenc} %
\usepackage[T1]{fontenc}    %
\usepackage{booktabs}       %
\usepackage{amsfonts}       %
\usepackage{nicefrac}       %
\usepackage{microtype}      %
\usepackage{arydshln}
\usepackage{mathtools,amssymb}
\usepackage{graphicx}
\usepackage{wrapfig}
\usepackage{multirow}
\usepackage{caption}
\usepackage[table]{xcolor}
\definecolor{trowcolor}{RGB}{213,231,230}
\definecolor{citecolor}{RGB}{91, 119, 147}
\definecolor{diffgreen}{RGB}{55,140,90}
\definecolor{diffred}{RGB}{180,85,85}
\hypersetup{
    colorlinks=true,
    citecolor=citecolor,
    linkcolor=black,
    urlcolor=citecolor
}
\title{Dex2HOI: Dexterous Bimanual Two-Object \\ Interaction Generation}

\author{
 \hspace{-0.55em} \textbf{Chrysa Pratikaki}$^{1}$ \quad
\textbf{Pablo Ruiz-Ponce}$^{2}$ \quad
\textbf{Jiankang Deng}$^{1}$ \\
\textbf{Stefanos Zafeiriou}$^{1}$ \quad
\textbf{Rolandos Alexandros Potamias}$^{1}$ \\[0.4em]
\normalfont $^{1}$Imperial College London, UK  \hspace{0.4em}
\normalfont $^{2}$University of Alicante, Spain
}

\iclrfinalcopy
\hypersetup{
  pdftitle={Dex2HOI: Dexterous Bimanual Two-Object Interaction Generation},
  pdfauthor={Chrysa Pratikaki, Pablo Ruiz-Ponce, Jiankang Deng, Stefanos Zafeiriou, Rolandos Alexandros Potamias}
}
\begin{document}

\maketitle

\begin{figure}[h]
  \vspace{-2em}
  \centering
  \includegraphics[width=1\linewidth]{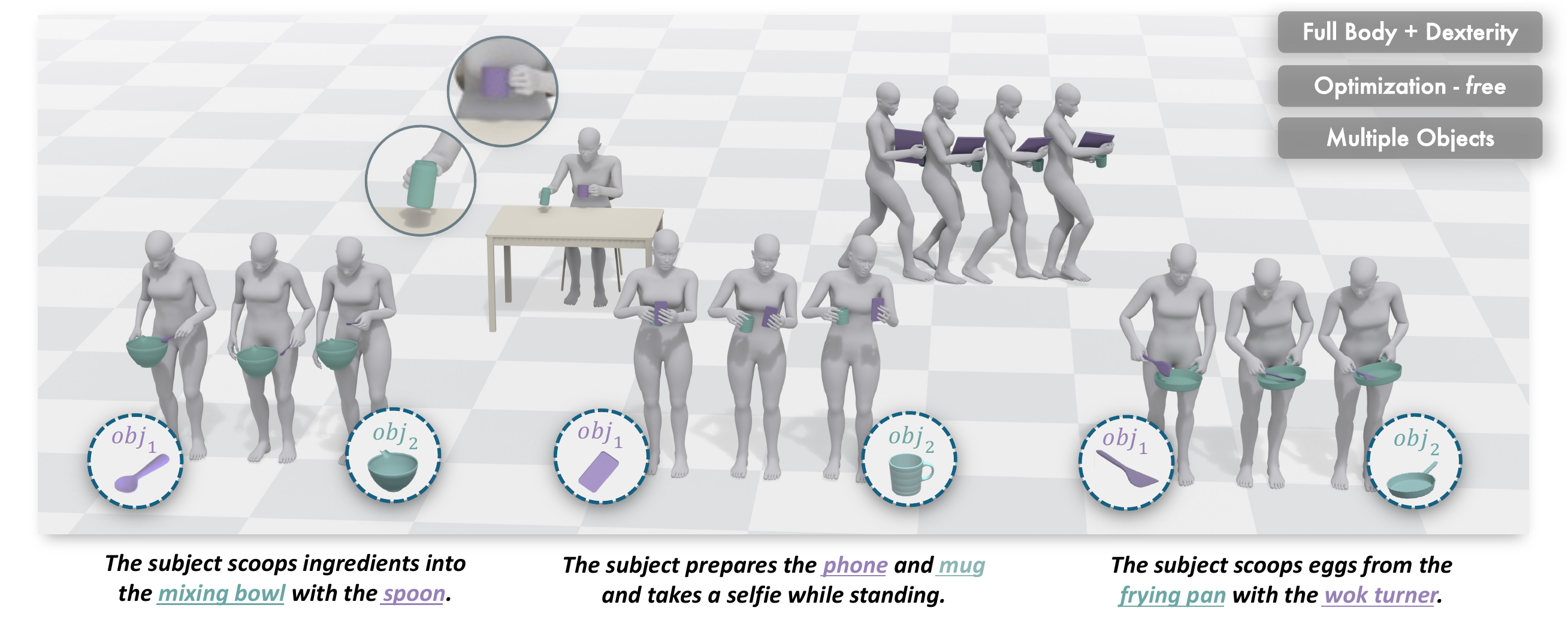}
\caption{ \textbf{Dex2HOI} is a diffusion-based model for Human-Object Interaction that generates dexterous bimanual manipulations from text, in single-shot, supporting interaction with up to two objects.}
  \label{fig:figteaser}
\end{figure}

\begin{abstract}
Recent advances in 4D Human-Object Interaction (HOI) generation have enabled increasingly realistic motion synthesis, particularly for single-object manipulation. Yet current research overlooks an inherent property of human behavior: people naturally coordinate both hands and manipulate multiple objects simultaneously. To address this gap, we present \textbf{Dex2HOI}, a unified diffusion model for single- and two-object HOI synthesis from text. At its core, Dex2HOI employs a Dual-Stream Diffusion approach, where each object is processed in a dedicated interaction stream and coordinated through bidirectional cross-attention. To synthesize the final motion, we introduce a Motion Fusion Network integrated with novel hand-relative object representations and contact-aware conditioning applied across the whole sequence. By sampling the diffusion process autoregressively over prefix-conditioned windows, Dex2HOI generates arbitrarily long sequences at real-time speed omitting redundant test-time optimization, achieving up to ×540 inference speed-up over prior state-of-the-art methods. Extensive evaluation on both single- and two-object benchmarks demonstrates \emph{state-of-the-art} quantitative results, marking a step beyond conventional single-object HOI generation and toward expressive multi-object manipulation. Project: \url{https://cpratikaki.github.io/dex2hoi/}
\end{abstract}
\vspace{-1em}
\section{Introduction}

Human motion is fundamentally shaped by interaction with our surroundings. We engage with objects in diverse ways, simultaneously manipulating multiple objects, or choosing to handle a single object with one or both hands depending on factors such as its weight, size, or the task at hand. This kind of coordinated bimanual multi-object manipulation, although fundamental to human-like behavior, remains severely underexplored in current human-object interaction (HOI) generation research ~\citep{Sui2026_hoisurvey}. Synthesizing such motions is more than a modelling curiosity: it underpins numerous applications, from realistic interactions in AR/VR systems, to embodied agents that assist in cluttered human environments and robotic systems that rely on manipulation priors ~\citep{chenobject_app1, pan2024synthesizing_app2, hoifhli}.

Recent progress in HOI generation has led to increasingly realistic motions on primarily single-object benchmarks, driven largely by diffusion-based models ~\citep{hoidiff2025, pi2025coda, ron2025hoidini, Zhang2025diffgrasp}. 
Despite this, most existing HOI methods face three limitations. First, nearly all state-of-the-art full-body methods assume only single-object interactions. Although a few datasets ~\citep{parahome_dataset, lv2024himonewbenchmarkfullbody, humoto_dataset} include multi-object sequences, dedicated generative methods remain largely unexplored. Second, many existing methods focus only on a subset of the human body: whole-body approaches such as the OMOMO ~\citep{li2023object_omomo} benchmark omit detailed hand articulation, while hand-centric methods like Text2HOI ~\citep{cha2024text2hoi} and LatentHOI ~\citep{Muchen_LatentHOI} partially address fine-grained grasping but sacrifice full-body generation. Third, current approaches in full-body HOI motion synthesis rely on auxiliary modules and optimization strategies to enforce accurate contacts. Some examples include predicting intermediate contact maps to guide generation ~\citep{diller2024cg_hoi}, conditioning the model on full object trajectories ~\citep{Zhang2025diffgrasp} or iterative test-time refinement such as DNO ~\citep{karunratanakul2023dno} with contact-aware constraints ~\citep{pi2025coda, ron2025hoidini}, resulting in inference overhead.

To address these limitations, we introduce a paradigm shift in both the representation of object motion and the architectural design of the generative model. We present \textbf{Dex2HOI}, a unified framework to tackle both single- and two-object full-body HOI synthesis from text. Our main objective is to approach HOI generation in a simple yet informed way: removing model redundancy, avoiding multi-stage pipelines, and producing accurate contacts in a single-shot diffusion sampling.
At the representation level, we model object motion as a learnable decomposition of hand-relative branches fused through learnable weights. This decomposition encodes a strong inductive bias for hand-object contact and naturally guides the diffusion process toward smooth transitions of control between hands.
At the architectural level, we propose the use of a Dual-Stream Diffusion decoder that processes each object in a dedicated interaction stream and models information across streams through bidirectional cross-attention. The body motions from each stream are then combined with our Motion Fusion Network to produce a single coherent motion. The same architecture handles \emph{both} one- and two-object sequences without modification, supporting a unified treatment of single- and bimanual manipulation.
Finally, we train an end-to-end model with contact-aware supervision, removing the need for test-time optimization or auxiliary contact-prediction modules. Together, these choices enable single-shot generation that is up to $\times$540 \textbf{faster} than prior optimization-based \emph{state-of-the-art} approaches. Overall, our contributions can be summarized as follows:

\begin{itemize}
  \item We propose the first \emph{unified} HOI generation model that synthesizes both single- and two-object interactions, advancing beyond existing single-object approaches.
    
    \item We introduce a novel hand-relative object motion representation and train end-to-end denoising models that enable efficient single-shot generation of complex sequences while maintaining physical plausibility and contact consistency across two hands and two objects.
    
    \item  We evaluate \textbf{Dex2HOI} generated motions across three benchmarks, GRAB ~\citep{taheri2020grab} for single-object HOI, and HIMO ~\citep{lv2024himonewbenchmarkfullbody} and HUMOTO ~\citep{humoto_dataset} for two-object HOI. We achieve state-of-the-art performance in both single- and two-object settings, which we further validate through VLM-based and user preference studies.
    
\end{itemize}

\section{Related Work}

\paragraph{Human Motion Generation.} Human motion generation is an open problem in computer vision, covering a broad set of tasks and applications. In recent years, diffusion-based models have emerged as the dominant paradigm ~\citep{chen2023executing_mld, mdm_tevet2023human, 10416192MotionDiffuse}, with the Motion Diffusion Model (MDM) ~\citep{mdm_tevet2023human} establishing a simple yet effective denoising framework for text-conditioned motion synthesis. Autoregressive formulations ~\citep{ar1,ar2} extend diffusion-based motion synthesis to long-horizon generation by predicting motion segments conditioned on a temporal context window, recursively using the latter frames as the condition for the next segment. Subsequent work extends these paradigms along several axes, beyond isolated body motion. Human-Human interaction methods ~\citep{intergen, ruiz2025mixermdm, interact2ar, InterControl} coordinate motion for multiple subjects, Human-Scene interaction methods ~\citep{cong2024laserhuman, scenediff, jiang2024autonomous, yi2024tesmo} condition motion in static 3D environments, while Human-Object interaction ~\citep{ghosh2022imos, hoidiff2025, pi2025coda, ron2025hoidini} targets interactions between the subject and usually one object at a time.

\paragraph{Human-Object Interaction Generation.} Progress in HOI generation has been particularly driven by the increasing presence of 4D HOI datasets, which span from small object manipulation datasets ~\citep{chao:cvpr2021DexYCB, taheri2020grab,  YangCVPR2022OakInk}, to interaction with larger everyday objects, including furniture and household devices ~\citep{chairs, li2023object_omomo, zhang2022behave} and some with even articulated objects ~\citep{fan2023arctic}. These datasets include interactions with a single object of interest at a time, while datasets that address the manipulation of more than two objects are more limited ~\citep{parahome_dataset, humoto_dataset, lv2024himonewbenchmarkfullbody}.

Early works on HOI generation ~\citep{ghosh2022imos, Taheri_2022_CVPR, grip_taheri2024} rely on conditional variational autoencoders (CVAEs) to model the distribution of plausible whole-body interactions. Many recent works have adopted denoising diffusion to improve realism, diversity, and conditioning flexibility in HOI synthesis ~\citep{diller2024cg_hoi, li2024chois, hoidiff2025, pi2025coda, ron2025hoidini,  wang2026unleashing, xu2025interact,  Zeng_2025_CVPR_chainhoi, Zhang2025diffgrasp, chen2026ho}. Among these, CoDA ~\citep{pi2025coda} and HOIDiNi ~\citep{ron2025hoidini} formulate HOI generation as a diffusion noise optimization problem, refining generated motion through contact-aware guidance to produce more plausible whole-body interactions, while CoDA further extends to articulated object manipulation. Additionally, DiffGrasp ~\citep{Zhang2025diffgrasp} chooses to generate whole-body grasping motions guided by whole sequence object trajectories. Other works focus on controllability and structure, including CHOIS~~\citep{li2024chois} for controllable interaction synthesis and Contact-Guided HOI~~\citep{diller2024cg_hoi} which leverages contact priors. Recent large-scale frameworks such as InterAct ~\citep{xu2025interact} further push toward generalizable single-object HOI animation.

Beyond purely generative formulations, a parallel line of work formulates HOI as a control problem and learns RL policies that produce physically plausible interactions ~\citep{braun2023physicallyplausiblefullbodyhandobject, wang2023physhoi, hoifhli, xu2025intermimic}. 
The aforementioned methods primarily focus on single-object interactions and do not explicitly address the generation of more complex two-object manipulations, leaving a significant gap in modeling coordinated bimanual interactions.

\paragraph{Hand-Object Interaction Generation.} A complementary line of research targets hand-centric interaction synthesis, focusing on generating fine-grained grasps. Early approaches such as ManipNet ~\citep{manipnet} predict dexterous finger motion using explicit hand-object spatial representations. More recent methods use denoising models to generate plausible hand-object interactions ~\citep{cha2024text2hoi, christen2024diffh2o, huang_etal_cvpr25, Muchen_LatentHOI,ye2023ghop,fu2026egograsp}. LatentHOI ~\citep{Muchen_LatentHOI} learns a latent hand diffusion model for generalizable grasp generation.  For more than one object, ManipTrans ~\citep{li2025maniptrans} addresses bimanual two-object grasping by transferring dexterous manipulation priors across objects via residual learning. While these methods achieve high-fidelity hand-object contact, they disregard full-body motion synthesis. Dex2HOI bridges this gap by jointly modeling dexterous bimanual contact and full-body motion within a single framework.

\section{Method}

\begin{figure}[t]
  \centering
  \includegraphics[width=1\linewidth]{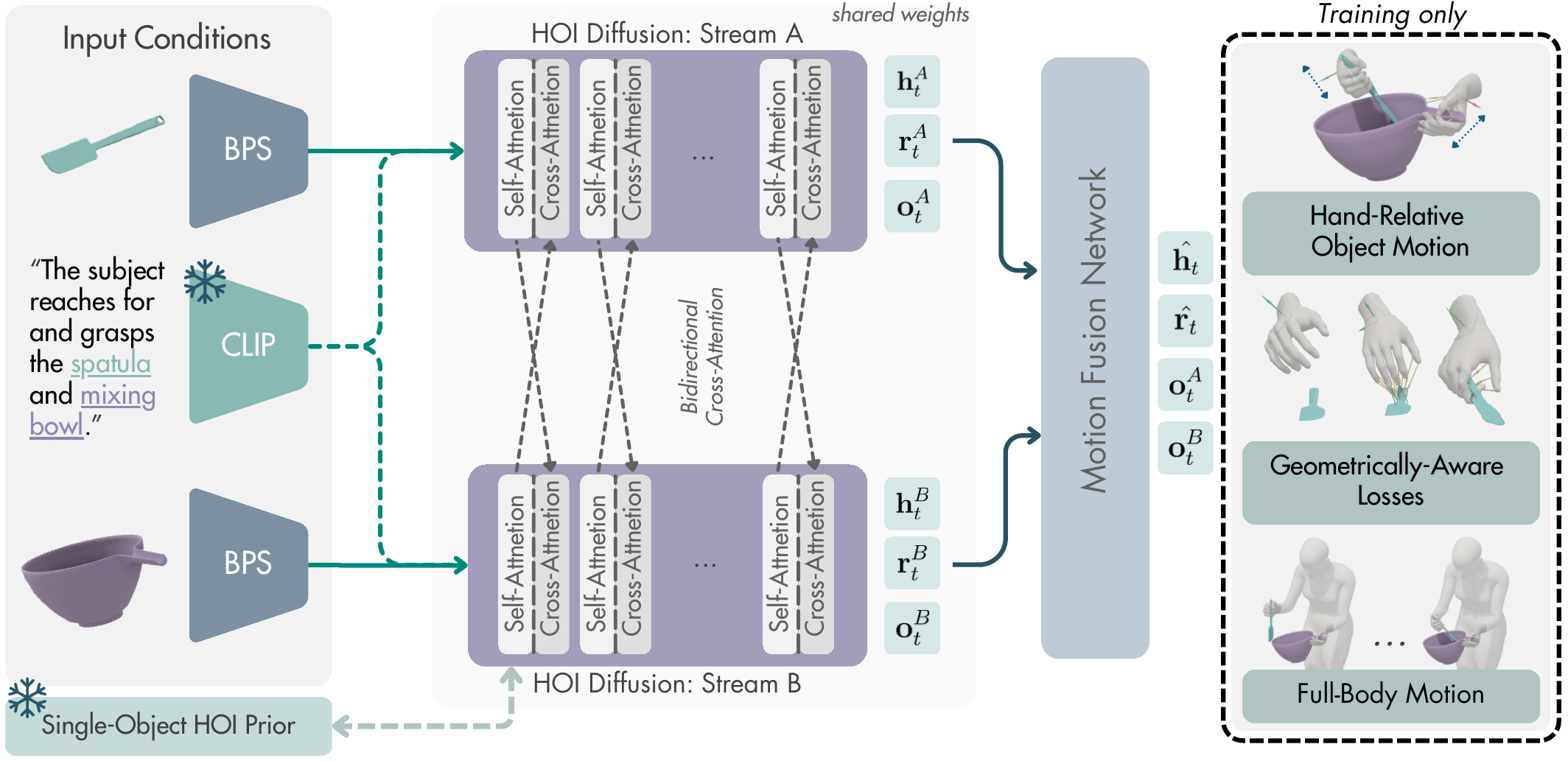}
  \caption{\textbf{Dex2HOI} generates dexterous bimanual two-object HOI sequences from text in single-shot, combining (a) Hand-Relative Object Motion Representations, (b) Dual-Stream HOI Diffusion, (c) our Motion Fusion Network, and (d) Geometrically-Aware Losses.}
  \label{fig:method}
\end{figure}

The overview of the proposed \textbf{Dex2HOI} is illustrated in Fig.~\ref{fig:method}. In Section~\ref{sec:representation} we introduce our \emph{hand-relative object representation}, which disentangles object motion into wrist-local and global branches. Section~\ref{sec:motionmodelling} then describes our Dual-Stream HOI diffusion and Motion Fusion Network that coordinate per-object streams into a single coherent full-body motion. Finally, in Section~\ref{sec:trainobjective} we introduce our geometry-informed interaction-aware training objective.

\subsection{Human and Object Motion Representations} \label{sec:representation}
We represent each motion sequence as a temporally aligned multimodal feature vector containing human rotations, human root translation and per-object trajectory representations. For a sequence of length $T$, the full motion feature at frame $t$ can be defined as:
\begin{equation}
\mathbf{f}_t = \left[ \mathbf{h}_t^{\{A,B\}},\; \mathbf{r}_t^{\{A,B\}},\; \mathbf{o}^{\{A,B\}}_t \right], \quad 0 \le t \le T
\end{equation}
where $\mathbf{h}_t$ denotes the human 6D ~\citep{Zhou_2019_CVPR_6drot} rotations, $\mathbf{r}_t \in \mathbb{R}^3$ denotes the human root translation, and $\mathbf{o}_t$ the \emph{hand-relative} object representations for the two objects ${\{A,B\}}$.

\paragraph{Human Representation.} We represent human body poses at frame $t$ as a set of 6D ~\citep{Zhou_2019_CVPR_6drot} joint rotations $\mathbf{h}_t \in \mathbb{R}^{6J}$, 
where $J$ denotes the number of joints. We set $J=52$ for datasets that utilize the SMPL-X parametric human model ~\citep{pavlakos2019expressive} and $J=65$ for datasets that utilize Mixamo parametric avatars~~\citep{mixamo}. 
To facilitate training, we also parameterize the 3D root translation $\mathbf{r}_t$ as the displacement relative to the first frame position (global pelvis bone).

\paragraph{Hand-Relative Object Representation.}  Modeling object motion in global coordinates alone is insufficient for predicting interaction-dependent trajectories: prior approaches consequently resort to multi-stage pipelines or test-time optimization to recover plausible contact. Relying purely on a global object trajectory provides a weak inductive bias and forces the network to implicitly infer the controlling hand at every frame. To address this, we introduce a \emph{hand-relative object representation} that explicitly disentangles object motion into right-hand, left-hand, and global components. Specifically, for frame $t$, we define the object feature vector as:
\begin{equation}
\mathbf{o}_t = \left[ \mathbf{r}^{obj}_t,\; \Delta^{L}_t,\; \Delta^{R}_t,\; \mathbf{p}^{G}_t,\; \boldsymbol{\alpha}_t \right] \in \mathbb{R}^{18},
\end{equation}
where $\mathbf{r}^{obj}_{t} \in \mathbb{R}^6$ denotes the global object rotation in 6D form ~\citep{Zhou_2019_CVPR_6drot},
$\Delta^{L}_t \in \mathbb{R}^3$ and $\Delta^{R}_t \in \mathbb{R}^3$ are the object translation expressed in the left and right wrist frames respectively, $\mathbf{p}^{G}_t \in \mathbb{R}^3$ represents the object translation in global coordinates and $\boldsymbol{\alpha}_t \in \mathbb{R}^3$ are learnable mixture weights that balance the contribution of wrist-relative displacements and global-coordinate translation.

Letting $\mathbf{p}^{L}_t,\mathbf{p}^{R}_t \in \mathbb{R}^3$ denote the global wrist positions, and $\mathbf{R}^{L}_t,\mathbf{R}^{R}_t \in SO(3)$ the corresponding global wrist rotations obtained from forward kinematics, we define the wrist-relative offsets, relative to the right ($R$) and left ($L$) hand: 
\begin{align}
\Delta^{\{R,L\}}_t &= (\mathbf{R}^{\{R,L\}}_t)^\top (\mathbf{p}^{obj}_t - \mathbf{p}^{\{R,L\}}_t)
\label{eq:deltas}
\end{align}
Intuitively, these terms describe where the object lies in each wrist's local frame. To ensure accurate object translation in cases where the object is not interacting with either hand, we include $\mathbf{p}^{G}_t$ as a global coordinate fallback. Under this formulation, we define the global object translation through a soft mixture of the three relative representations as: 
\begin{equation}
\hat{\mathbf{p}}^{obj}_t
=
w^L_t \left(\mathbf{p}^{L}_t + \mathbf{R}^{L}_t \Delta^{L}_t \right)
+
w^R_t \left(\mathbf{p}^{R}_t + \mathbf{R}^{R}_t \Delta^{R}_t \right)
+
w^G_t \mathbf{p}^{G}_t,
\label{eq:trajdecode}
\end{equation}
where $[w^L_t,w^R_t,w^G_t] = \mathrm{softmax}(\boldsymbol{\alpha}_t)$ are the learned mixing weights. 

Using this formulation, the model learns to smoothly interpolate between left-hand control, right-hand control, and global motion through the learned weights $[w^L_t, w^R_t, w^G_t]$, which adapt at each frame depending on which hand is in contact with the object, instead of forcing all interaction regimes into a single translation branch.

\subsection{Motion Modeling}\label{sec:motionmodelling}
We formulate human-object motion synthesis as autoregressive conditional denoising over structured motion sequences, enabling the generation of arbitrarily long sequences ~\citep{ron2025hoidini, tevet2025closd}. Given an observed prefix of length $T_p$, the AR model $D_{\theta}$ predicts the future segment $\mathbf{F}_{T_p:T_p+T_f-1}$, with a prediction length of $T_f$ frames.

\paragraph{Dual-Stream HOI Diffusion $\mathbf{D}_{\mathbf{\theta}}$.} To facilitate motion generation for multiple objects, we adopt a cooperative denoising paradigm ~\citep{intergen} with two coordinated diffusion streams $A$ and $B$ with shared weights, one for each object-conditioned interaction. Stream $A$ carries motion representation  $\mathbf{f}^A_t = [\mathbf{h}_t,\,\mathbf{r}_t,\,\mathbf{o}^{A}_t]$ and
stream $B$ carries $\mathbf{f}^B_t =[\mathbf{h}_t,\,\mathbf{r}_t,\,\mathbf{o}^{B}_t]$,
where both share the human representation $[\mathbf{h}_t, \mathbf{r}_t]$ but differ in their object representation. At diffusion timestep $\tau$, the shared
denoiser $D_\theta$ processes each stream conditioned on the other:
\begin{equation}
  \hat{\mathbf{f}}^A_0
  = D_{\theta}\!\left(\mathbf{f}^A_\tau,\,\mathbf{f}^B_\tau,\,\mathbf{c}^A,\,\tau\right),
  \qquad
  \hat{\mathbf{f}}^B_0
  = D_{\theta}\!\left(\mathbf{f}^B_\tau,\,\mathbf{f}^A_\tau,\,\mathbf{c}^B,\,\tau\right),
  \label{eq:denoise}
\end{equation}
where $\mathbf{c}^A$ and $\mathbf{c}^B$ are branch-specific conditioning
signals containing: a frozen CLIP ~\citep{clip} text embedding; object
geometry encoded as Basis Point Set (BPS) features
~\citep{prokudin2019efficient}; and a lightweight MLP projection for $\tau$. To enable information flow between the two HOI streams, we follow ~\citep{intergen} and apply bidirectional cross-attention $\mathrm{CA}(\cdot)$ after each layer $\ell$ of the shared HOI Diffusion $D_{\theta}$, resulting in each stream effectively attending to the other.

\paragraph{Motion Fusion Network $\mathbf{D}_{\mathbf{h}}$.}At the output of $D_{\theta}$, both streams produce a complete prediction $\hat{\mathbf{f}}^{A}_0$ and $\hat{\mathbf{f}}^{B}_0$, each containing full motion features with human representation and hand-relative object representation per object present. We decompose each prediction along the feature partition introduced in Section~\ref{sec:representation}:
\begin{equation}
\hat{\mathbf{f}}^{A}_0 = \big[\, \hat{\mathbf{h}}^A,\; \hat{\mathbf{r}}^A,\; \hat{\mathbf{o}}^{A} \,\big],
\qquad
\hat{\mathbf{f}}^{B}_0 = \big[\, \hat{\mathbf{h}}^B,\; \hat{\mathbf{r}}^B,\; \hat{\mathbf{o}}^{B} \,\big].
\end{equation}
To produce a single coherent human pose, derived from the separate HOI Diffusion streams $[\hat{\mathbf{h}}^A,\hat{\mathbf{r}}^A]$ and $[\hat{\mathbf{h}}^B,\hat{\mathbf{r}}^B]$, we introduce a dedicated \emph{Motion Fusion Network} $D_{\mathbf{h}}$, implemented as a stack of self-attention layers that operates \emph{only} on the human representations (Figure~\ref{fig:method}):
\begin{equation}
  \big[\hat{\mathbf{h}},\; \hat{\mathbf{r}}\big] = D_{\mathbf{h}}\!\left(\big[\hat{\mathbf{h}}^A,\hat{\mathbf{r}}^A\big],\; \big[\hat{\mathbf{h}}^B,\hat{\mathbf{r}}^B\big]\right).
\end{equation}
The fused body prediction $[\hat{\mathbf{h}}, \hat{\mathbf{r}}]$ replaces the per-stream body outputs, while the object predictions $\hat{\mathbf{o}}^{A}$ and $\hat{\mathbf{o}}^{B}$ are taken directly from their originating streams without modification. This design enforces a clean factorization of the generation process: each object's dynamics are denoised under its own object-specific conditioning, while the body is reconciled in a shared latent that accesses both interactions. As we demonstrate through ablations in Section~\ref{sec:baselines}, the Motion Fusion Network outperforms simpler alternatives such as averaging or selecting one of the two human predictions, which often lead to inconsistent motions.

\subsection{Training}\label{sec:trainobjective}

Since our HOI diffusion model solely predicts rotation parameters for the human body, we apply a differentiable forward kinematics layer $\mathrm{FK}(\cdot)$ to map predicted poses to world-space joint positions, thus enabling velocity and contact losses that are otherwise inaccessible in the rotation domain.
Along with the main diffusion objective $\mathcal{L}_{diff}$ and the well-established geometry regularization and SDF losses~\citep{intergen, pi2025coda, ron2025hoidini}, which cover joint and position velocity (supervision against target), and interpenetrations (body-object and inter-object SDFs),  we also enhance our training objective by introducing the following contact-aware conditioning:

\paragraph{Hand-relative object representation supervision.} Our hand-relative object representation $\mathbf{o}_t$ induces a structured decomposition of object motion into wrist-relative and global branches, according to Eq.~\ref{eq:trajdecode}. We explicitly supervise the different components. First, we supervise the mixture weights by comparing the predicted weights $[\hat{w}^L_t,\hat{w}^R_t,\hat{w}^G_t] = \mathrm{softmax}(\hat{\boldsymbol{\alpha}}_t)$, extracted from the predicted representation $\hat{\mathbf{o}}t$, against hand proximity-derived ground-truth targets $[w^L_t,w^R_t,w^G_t]$: 

\begin{equation}
\mathcal{L}_{mix}
=
\left\|\hat{w}^{
s}_t - w^{s}_t
\right\|_2^2, \quad s \in \{L, R, G\}.
\end{equation}
Second, we supervise the decoded object translation $\hat{\mathbf{p}}^{obj}_t$, obtained via the mixture decoder of Eq.~\ref{eq:trajdecode}, against the ground-truth world-space position $\mathbf{p}^{obj}_t$: 
\begin{equation} \mathcal{L}_{trans}=
\left\| \hat{\mathbf{p}}^{obj}_t - \mathbf{p}^{obj}_t \right\|_2^2, \end{equation} 
enforcing correct trajectory reconstruction while ensuring gradients propagate through all branches and mixture weights. Finally, we regularize the predicted wrist-relative offsets $\hat{\Delta}^{L}_t$ and $\hat{\Delta}^{R}_t$, as defined in Eq.~\ref{eq:deltas}, against the ground-truth ${\Delta}^{L}_t$ and ${\Delta}^{R}_t$, further encouraging geometric consistency: 
\begin{equation} \mathcal{L}_{off}=
\frac{1}{2}
\left(
\left\|\hat{\Delta}^{L}_t - \Delta^{L}_t\right\|_2^2
+
\left\|\hat{\Delta}^{R}_t - \Delta^{R}_t\right\|_2^2
\right).
\end{equation}

\paragraph{Hand-Object contact regression.}
We enforce fine-grained interaction geometry via a unified finger-object distance loss. Letting $d_{t,j}$ denote the
ground-truth signed nearest-surface distance from hand  joint $j$ at frame $t$ to the object mesh, and $\hat{d}_{t,j}$ the predicted counterpart,  we regress nearest-surface distances collectively via Huber loss $\rho_\delta$:
 \begin{equation} \mathcal{L}_{dist}=
\frac{1}{|\mathcal{A}|}
\sum_{(t,j)\in\mathcal{A}}
\rho_\delta\left(\hat{d}_{t,j} - d_{t,j}\right),
\end{equation}
where $\mathcal{A} = {(t,j) : d_{t,j} < d_{min}}$ selects %
frames where 
hand joint $j$ lies within approach radius $d_{min}=0.10$ of the object surface. 
The full training objective is additionally discussed in the Appendix.

\section{Evaluation}

\subsection{Datasets}\label{sec:datasets}
To train and evaluate Dex2HOI we use the following datasets:

\textbf{GRAB.} A dataset~\citep{taheri2020grab} with 10 subjects interacting with 51 everyday objects. We follow the typical split ~\citep{ron2025hoidini, pi2025coda} using subjects 1 through 9 for training and subject 10 (145 sequences) as the test set.

\textbf{HUMOTO.} A mo-cap dataset~\citep{humoto_dataset} that contains both single and multi-object manipulation tasks in Mixamo representations~\citep{mixamo}. We retain clips that contain up to two moving objects, splitting according to the provided annotations at a frame level. The resulting benchmark yields 337 train and 45 test sequences, mostly from tabletop manipulation scenes; the broader 1/2-object subset used for one-object training and evaluator training contains 1828 train and 322 test sequences.

\textbf{HIMO.} A multi-object interaction dataset ~\citep{lv2024himonewbenchmarkfullbody} covering simultaneous manipulation of 2 or 3 objects. The 2-object split contains 2578 train and 522 test sequences, and the 3-object split contains 1045 train and 176 test sequences, following the official splits. In this work, we use the 2-object split to evaluate our work. More about the datasets can be found in the Appendix.  

\subsection{Motion Evaluators and Metrics}\label{sec:metrics}

To evaluate the generated motions on the HUMOTO interactions, we trained a dedicated transformer-based motion evaluator on ground-truth motions of the human joints and of the two objects, following the protocol in~\citep{ghosh2022imos}. The motion encoder is used to compute FID$\downarrow$, Diversity$\rightarrow$, and MultiModality (MM $\uparrow$ / MModality $\uparrow$) based on extracted motion embeddings and assess distributional quality, while for text-alignment metrics MultiModal Distance (MMdist $\downarrow$) and R-Precision (R@3 $\uparrow$), we train an additional motion-text aligner using a contrastive objective. Finally, the penetration depth metric measures the mean hand and body–object interpenetration distance across frames. 

\begin{figure}[t]
  \centering
 
  \includegraphics[width=0.95\linewidth]{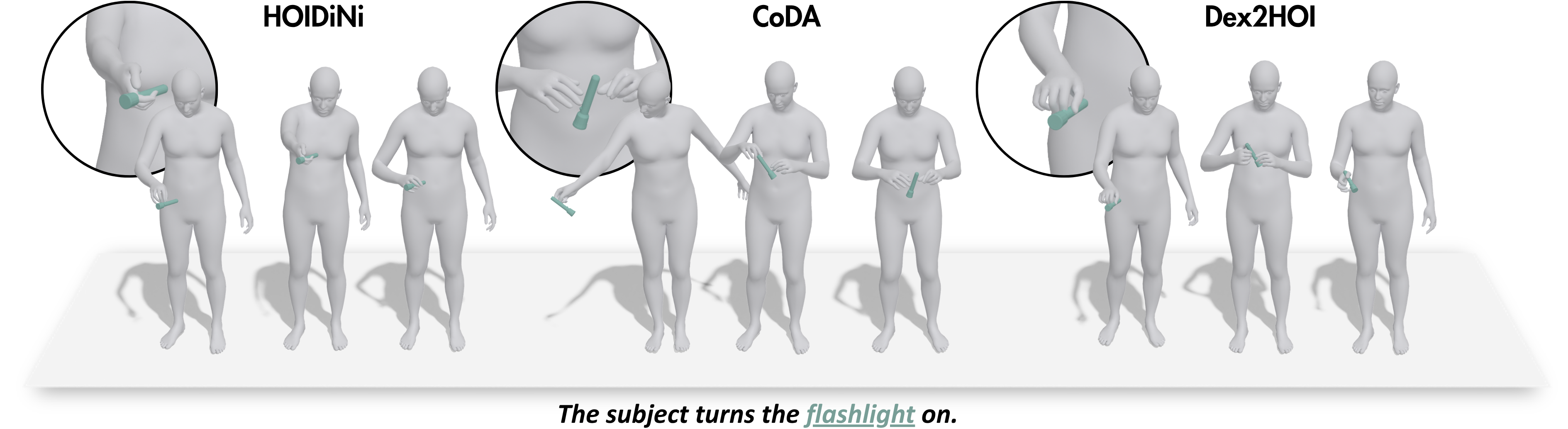}
  \caption{\textbf{Dex2HOI} comparative qualitative results on GRAB (single-object HOI synthesis).}
  \label{fig:grabqual}
\end{figure}

\begin{table}[t]
\centering

\resizebox{\linewidth}{!}{%
\begin{tabular}{l|lcccccc|c}
\toprule
 & Method & FID $\downarrow$ & Diversity $\rightarrow$ & MM $\rightarrow$ & Pen.\ (mm) $\downarrow$ & R@3 $\uparrow$ & MMdist $\downarrow$ & Runtime (s) $\downarrow$ \\
\midrule
\multirow{9}{*}{\rotatebox{90}{1 object }}

& GT                   & ---    & 0.706 & 0.194    & ---    & 0.617 & --- & --- \\
\cdashline{2-9}\noalign{\vskip 2pt}
& MDM             & 0.514 & 0.674 & 0.186 &    5.85 & {0.570} & \underline{1.137} & 0.56{\scriptsize$\pm$0.03} \\
& IMoS  (w/o obj optim)            & {0.521} & 0.608 & 0.207 & {6.23} & 0.557 & 1.165 & 0.75{\scriptsize$\pm$0.05} \\
& IMoS               & \underline{0.494} & 0.632 & \underline{0.189} & {7.02} & 0.557 & 1.165 & 59.22{\scriptsize$\pm$2.15} \\
& HOIDiNi  (w/o DNO)     & 0.503 & \underline{0.715} & 0.208 & 10.88 & {0.587} & \textbf{1.126} & \underline{0.43}{\scriptsize$\pm$0.02} \\
& HOIDiNi    & 0.584 & 0.783 & 0.183 &  7.34 & 0.528 & 1.179 & 122.70{\scriptsize$\pm$3.41} \\
& CoDA  (w/o DNO)        & 0.697 & 0.483 & \textbf{0.189} &  7.39 & 0.567 & 1.369 & 2.79{\scriptsize$\pm$0.08} \\
& CoDA        & 0.558 & 0.694 & 0.249 & \textbf{5.36} & \underline{0.592} & 1.446 & 135.26{\scriptsize$\pm$12.7} \\

\rowcolor{trowcolor}
\cellcolor{white} & \textbf{Dex2HOI}  & \textbf{0.479} & \textbf{0.702} & 0.161 & \underline{5.70} & \textbf{0.593} & 1.222 & \textbf{0.25}{\scriptsize$\pm$0.01} \\
\bottomrule
\end{tabular}}
 \captionsetup{belowskip=-8pt}
\caption{Single-object motion evaluation on the GRAB dataset ~\citep{taheri2020grab}: Dex2HOI provides a \emph{favorable} trade-off across realism, multimodality and action retrieval metrics, while maintaining quality hand-object contacts and without utilizing test-time optimization ~\citep{karunratanakul2023dno}. In the last column we report the average time it takes to infer a 60-frame sample. We highlight \textbf{best} and \underline{second best} results and ($\rightarrow$) indicates results closer to the ground-truth values are better. Results across all tables are averaged over 10 independent seeded runs.}
\label{tab:grabeval}
\end{table}

To evaluate on GRAB we follow the existing protocol~\citep{ghosh2022imos,ron2025hoidini}. In contrast to the HUMOTO evaluator that is trained using a contrastive learning objective, the GRAB evaluator serves as an action-recognition network, since GRAB comes with discrete action labels~\citep{taheri2020grab}. We report the same six metrics across all GRAB experiments, with the difference that MultiModality is calculated between generations of the same \emph{action} labels instead of instruction prompts, following \citep{ghosh2022imos}. 

\subsection{Experiments}\label{sec:baselines}

\begin{wrapfigure}[18]{r}{0.3\linewidth}
  \centering
  \vspace{-1.7em}
  \includegraphics[width=\linewidth]{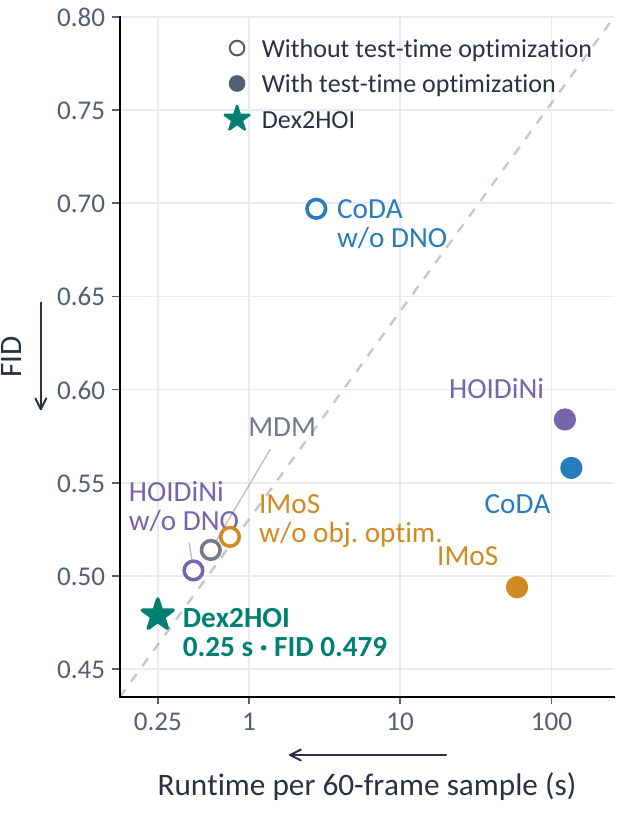}
  \caption{Optimization Pareto Diagram - lower FID and faster runtime is better.}
  \label{fig:pareto}
\end{wrapfigure}
\paragraph{Baselines.} On the two-object benchmarks HUMOTO and HIMO, we compare \textbf{Dex2HOI} against the HIMO-Gen generative baseline~\citep{lv2024himonewbenchmarkfullbody} and an MDM baseline~\citep{mdm_tevet2023human}, adapted to jointly predict object and human motion. Given the limited availability of multi-object evaluation benchmarks and to also evaluate our separate HOI Diffusion Models $D_{\theta}$, we perform additional evaluation on the single-object against previous state-of-the-art full-body HOI methods, namely HOIDiNi~\citep{ron2025hoidini}, CoDA~\citep{pi2025coda} and IMoS~\citep{ghosh2022imos} and MDM~\citep{mdm_tevet2023human}.  
Apart from MDM, these methods employ some form of hand-object contact informed optimization at inference, producing plausible results at an additional computational cost. Adapting these optimization frameworks to two-object sequences is a non-trivial task; Their contact objectives are formulated around a single object–hand pair, with no mechanism to resolve which hand should contact which object at each timestep. Naively running per-object optimization in parallel without coordination signals risks yielding inter-object penetrations and incoherent motion.

\begin{figure}[t]
  \centering
  \includegraphics[width=0.95\linewidth]{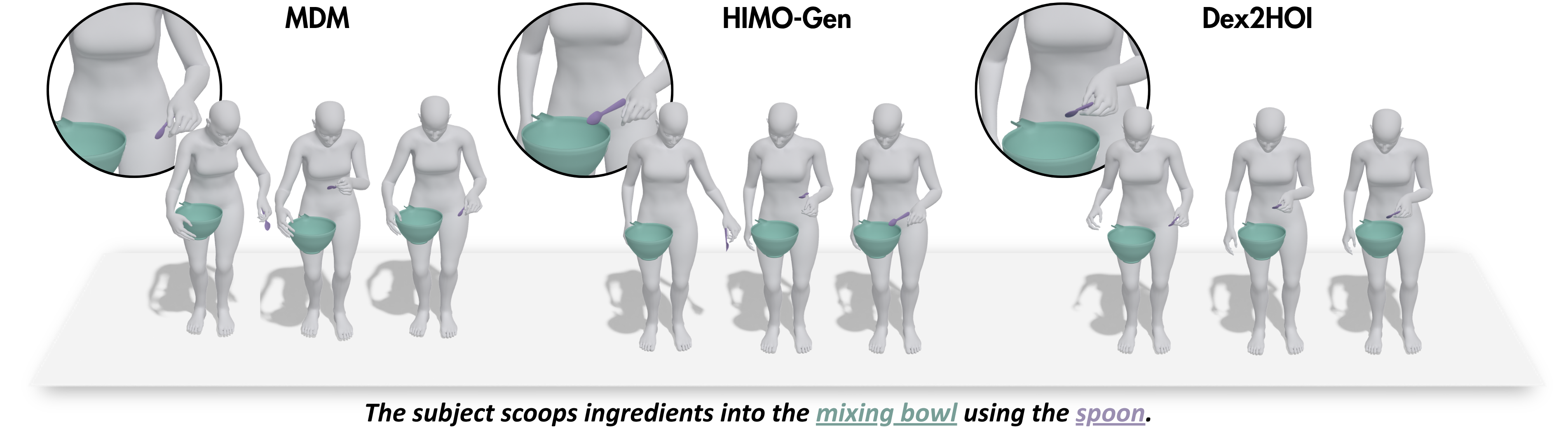}
  \caption{Qualitative contact comparison against HIMO-Gen and MDM on HUMOTO.}
  \label{fig:humotoqual}
\end{figure}

\begin{table}[t]
\centering

\resizebox{\linewidth}{!}{%
\begin{tabular}{l|l c c c c c c}
\toprule
 & Method & FID $\downarrow$ & Diversity $\rightarrow$ & MModality $\uparrow$ & Pen. (mm) $\downarrow$ & R-Pre@3 $\uparrow$ & MMdist $\downarrow$ \\
\midrule
& GT & --- & 1.061 & --- & --- & 0.520 & 1.210 \\
\cdashline{2-8}\noalign{\vskip 2pt}
& MDM ~\citep{mdm_tevet2023human} & 0.896 & \textbf{0.797} & 0.009 & 12.8 & 0.118 & 1.420 \\
& HIMO-Gen ~\citep{lv2024himonewbenchmarkfullbody} & \underline{0.886} & 0.606 & \underline{0.019} & \underline{11.7} & \textbf{0.295} & \underline{1.281} \\
\rowcolor{trowcolor}
\multirow{-4}{*}{\rotatebox{90}{2 objects}} \cellcolor{white} & \textbf{Dex2HOI} & \textbf{0.655} & \underline{0.780}& \textbf{0.056} & \textbf{9.8} & \underline{0.275} & \textbf{1.260} \\
\midrule
& GT & --- & 1.192 & --- & --- & 0.665 & 1.250 \\
\cdashline{2-8}\noalign{\vskip 2pt}
& MDM ~\citep{mdm_tevet2023human} & \underline{1.394} & \underline{0.464} & \underline{0.053} & \underline{18.2} & 0.086 & 1.445 \\
& HIMO-Gen ~\citep{lv2024himonewbenchmarkfullbody} & 1.418 & 0.413 & 0.008 & 20.7 & \underline{0.188} & \underline{1.410} \\
\rowcolor{trowcolor}
\multirow{-4}{*}{\rotatebox{90}{1 object}} \cellcolor{white} & \textbf{Dex2HOI} & \textbf{0.880} & \textbf{0.848} &\textbf{ 0.103} & \textbf{13.1} & \textbf{0.250} & \textbf{1.323} \\
\bottomrule
\end{tabular}}
 \captionsetup{belowskip=-8pt}
\caption{\textbf{Quantitative evaluation on the HUMOTO dataset~\citep{humoto_dataset}}, reporting performance on both two-object (top) and single-object (bottom) settings. We compare \textbf{Dex2HOI} against HIMO-Gen~\citep{lv2024himonewbenchmarkfullbody} and MDM~\citep{mdm_tevet2023human} and highlight \textbf{best} and \underline{second-best} results.}\label{tab:humotoeval}
\end{table}

\paragraph{Quantitative Results.} As shown in Table~\ref{tab:grabeval}, \textbf{Dex2HOI} achieves the best FID, Diversity and R@3 on GRAB while running up to $\times$540 times
faster on average than CoDA and $\times$490 times than HOIDiNi, depending on the selected duration of the iterative noise optimization (we use 200 opt. steps on both methods over 60-frame predictions). Notably, our single-shot generation surpasses optimization-based methods across standardized metrics, without the exhaustive runtime requirements - which is also summarized in the Pareto analysis in Figure~\ref{fig:pareto}. Similarly, on the HUMOTO dataset (Table~\ref{tab:humotoeval}), \textbf{Dex2HOI} achieves the best FID, penetration, and MModality across both single- and two-object settings, with a great gain over prior methods particularly on single-object instances.
\paragraph{Ablation Study.}
\begin{wrapfigure}[17]{r}{0.4\linewidth}
  \centering
  \includegraphics[width=\linewidth]{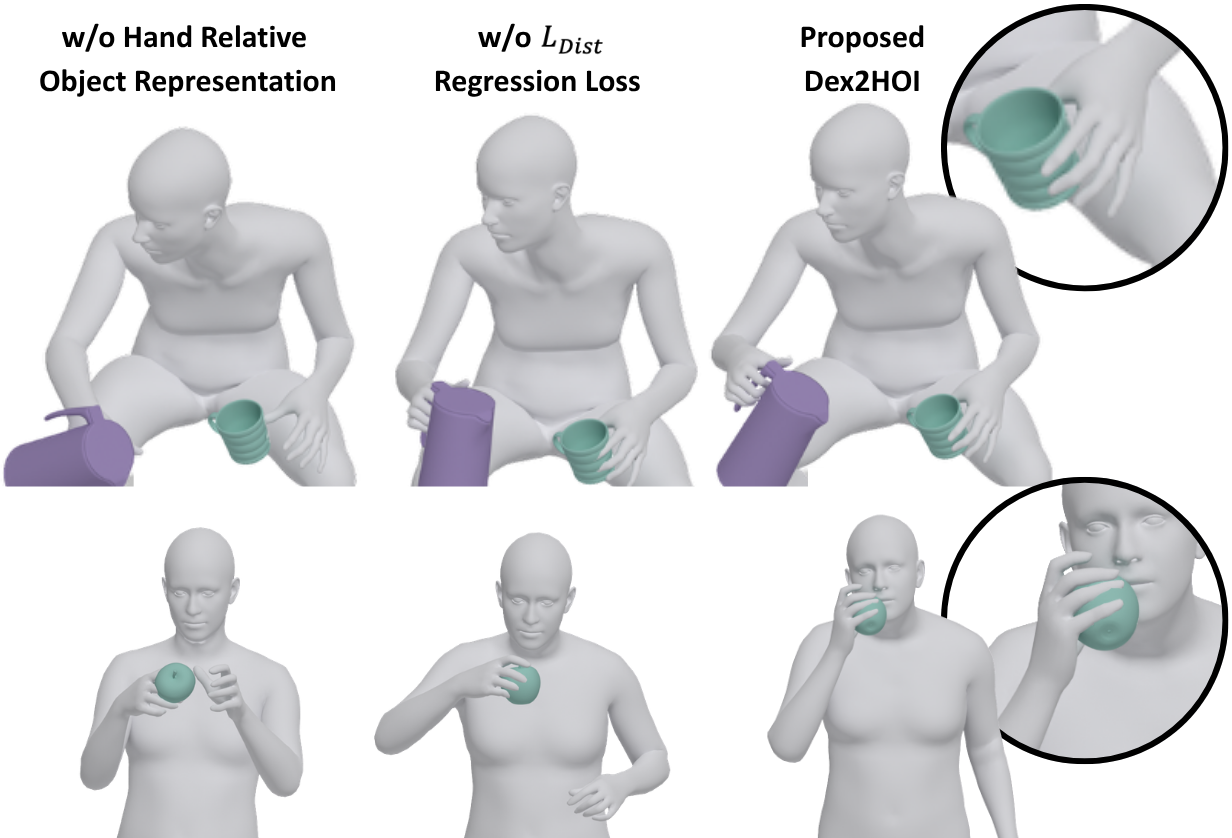}
\caption{\textbf{Representation and contact ablations.} Hand-relative representations couple object motion to the controlling wrists, while $L_{dist}$ supervises fine-grained hand-object contact.}
  \label{fig:abl}
\end{wrapfigure}
We design our ablation study around the core components of Dex2HOI and we provide the following variants for two-object evaluation: \textbf{(i) w/o the Motion Fusion Network}, where we replace the learned $D_{\mathbf{h}}(\cdot)$ model with an element-wise average of the two stream body predictions.  \textbf{(ii) w/o Dual-stream HOI}, where we use a single denoising stream that jointly predicts human motion together with both objects. \textbf{(iii) w/o hand-relative object representations}, where we replace the proposed representation with global object trajectory prediction. \textbf{(iv) w/o mixture supervision}, where we remove the $\mathcal{L}_{\text{mix}}$, $\mathcal{L}_{\text{trans}}$, and $\mathcal{L}_{\text{off}}$ terms, leaving only the diffusion MSE loss on the hand-relative object representation. This tests whether the hand-relative representation can be learned implicitly, or whether weight supervision is required.   \textbf{(v) w/o finger-contact supervision $\mathcal{L}_{\text{dist}}$}, where we remove our contact regression loss from the learning objective. Finally, \textbf{(vi) GRAB single prior}, where the diffusion layers are initialized from frozen single-object GRAB training, and only cross-attention and \(D_h\) are learned from two-object data. On GRAB, we further ablate the following: (i) hand-relative representations, (ii) mixture supervision, and (iii) \(L_{dist}\).

\paragraph{Ablation Results.} Table~\ref{tab:ablation} shows that the proposed gains do not arise from simply scaling a motion diffusion model to additional output dimensions. Replacing the hand-relative representation with global object trajectories causes the largest degradation, while removing the dual-stream formulation or learned Motion Fusion consistently reduces generation quality, supporting explicit object-wise modeling and learned coordination. Mixture supervision further improves both motion and interaction quality, indicating that explicitly modeling transitions between left-hand, right-hand, and global object control is beneficial. Notably, removing \(L_{dist}\) leaves FID nearly unchanged (\(0.655\!\rightarrow\!0.656\)) but increases penetration from \(9.8\) to \(11.6\) mm, showing that it specifically improves local interaction geometry rather than global motion statistics. Qualitative results are also shown in Figures \ref{fig:grabqual} and \ref{fig:humotoqual} for GRAB and HUMOTO respectively, while Figure~\ref{fig:abl} qualitatively illustrates the effect of the proposed representation and contact supervision.  Finally, although \textbf{Dex2HOI} is designed for single-shot generation, we additionally provide an \textbf{iterative refinement} ablation for applications that prioritize contact fidelity over inference speed. We use a deterministic sequence-level training-free optimizer which solves for a gated per-frame correction to the elbow, wrist and finger rotations of each hand under a 5 mm contact gate. The refinement results in minor finger-distance corrections while leaving the body, root trajectory, and object trajectory unchanged. This procedure is strictly post-hoc and not used in \textbf{any visual result in this paper}, unless otherwise specified in the Appendix. 

\begin{table}[h]
\centering
\resizebox{\linewidth}{!}{%
\begin{tabular}{l|lcccccc}
\toprule
& Method & FID $\downarrow$ & Diversity $\rightarrow$ & MModality $\uparrow$ & Pen. (mm) $\downarrow$ & R-Pre@3 $\uparrow$ & MMdist $\downarrow$ \\
\midrule
& GT & --- & 1.061 & --- & --- & 0.520 & 1.210 \\
\cdashline{2-8}\noalign{\vskip 2pt}
& w/o Fusion (avg layer) & 0.758 & 0.761 & 0.055 & \underline{10.9} & 0.271 & \underline{1.255} \\
& w/o Dual-stream HOI & 0.885 & 0.638 & 0.025 & 11.4 & 0.243 &\textbf{ 1.255} \\

& w/o hand-relative obj reps & 0.950 & 0.514 & 0.026 & 13.8 & 0.262 & 1.281 \\
& w/o mixture supervision & 0.831 & 0.586 & 0.033 & 11.8 & 0.167 & 1.294 \\
& w/o contact regression $L_{Dist}$ & 0.656 & 0.749 & 0.063 & 11.6 & 0.274 & 1.260 \\
& w/ GRAB single prior & \textbf{0.576} & \underline{0.766} & \textbf{0.109} & 11.4 & \underline{0.275} & 1.259 \\
\rowcolor{trowcolor}
\cellcolor{white} & \textbf{Proposed} & \underline{0.655} & \textbf{0.780}& \underline{0.056} & \textbf{9.8} & \textbf{0.275} & {1.260} \\
\cdashline{2-8}\noalign{\vskip 2pt}
 \rowcolor{trowcolor!30!white}
\cellcolor{white}\multirow{-8}{*}{\rotatebox{90}{2 objects}} &
\quad + contact refinement ($\Delta$) &
\textcolor{diffgreen}{\textit{-0.001}} &
\textcolor{diffgreen}{\textit{+0.012}} &
\textit{+0.000} &
\textcolor{diffgreen}{\textit{-0.3}} &
\textit{+0.000} &
\textit{-0.000} \\
\midrule
& GT                   & ---    & 0.706 & 0.194    & ---    & 0.617 & 1.175  \\
\cdashline{2-8}\noalign{\vskip 2pt}
& w/o hand-relative obj reps & 0.621 & 0.651 & 0.149 & 10.9 & 0.437 & 1.346 \\
& w/o mixture supervision & 0.596 & 0.677 & \underline{0.159} & 9.9 & \underline{0.466} & \underline{1.261} \\
& w/o contact regression $L_{Dist}$ & \underline{0.570} & \underline{0.679} & 0.154 & \underline{6.6} & 0.421 & 1.374 \\
\rowcolor{trowcolor}
\cellcolor{white} & \textbf{Proposed} & \textbf{0.479} & \textbf{0.702} & \textbf{0.161} & \textbf{5.7} & \textbf{0.593} & \textbf{1.222}  \\
\cdashline{2-8}\noalign{\vskip 2pt}
\rowcolor{trowcolor!30!white}
\cellcolor{white}\multirow{-6}{*}{\rotatebox{90}{1 object}} &
\quad + contact refinement ($\Delta$) &
{\textit{-0.000}} &
\textcolor{diffgreen}{\textit{+0.011}} &
\textit{+0.000} &
\textcolor{diffgreen}{\textit{-1.3}} &
\textit{+0.000} &
\textcolor{diffred}{\textit{+0.005}}\\
\bottomrule
\end{tabular}}
\caption{\textbf{Ablation study on HUMOTO (2 objects) and GRAB (1 object).} Removing the Hand-Relative object representations and other key model parts consistently degrades performance, confirming that both components are essential. The GRAB-pretrained variant remains competitive, indicating that grasping priors transfer effectively from SMPL-X (GRAB) to Mixamo characters.}
\label{tab:ablation}
\end{table}

\begin{figure}[h]
  \centering
  \includegraphics[width=0.9\linewidth]{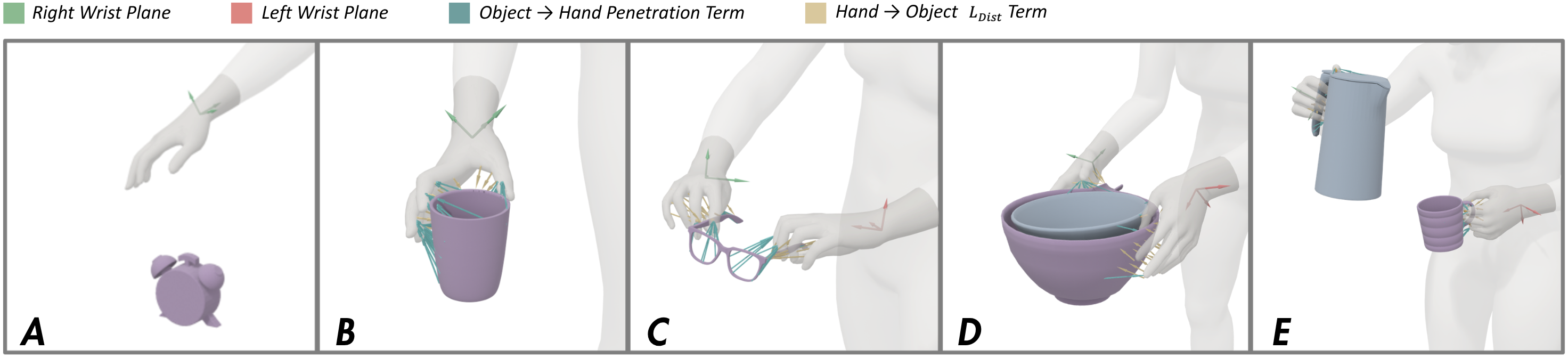}
  \caption{Dex2HOI's hand-relative object representation models object translation relative to the wrists while retaining a global object component, allowing our unified model to represent diverse hand–object interactions. \textbf{(A) No-contact}: the object remains spatially separated from both hands, using global object motion fallback. \textbf{(B) Single-hand contact:} the object is primarily constrained relative to a single wrist, supporting one-handed grasping. \textbf{(C) Bimanual handling} of a single object and \textbf{(D)} of multiple objects, and \textbf{(E) Independent two-object manipulation}.}
  \label{fig:handrepimage}
\end{figure}

\paragraph{Perceptual Assessment.} We complement metric-based evaluation with VLM and human perceptual studies assessing three properties central to dexterous HOI:  motion naturalness, hand–object contact quality and text prompt alignment. For the VLM evaluation, we query InternVL3-8B to evaluate rendered while, separately, participants compare anonymized side-by-side renderings of competing methods. Full protocols, prompts, and results are provided in the Appendix.

\section{Conclusion}\label{sec:conclusion}

We presented \textbf{Dex2HOI}, a unified framework for human-object interaction generation that extends beyond conventional single-object HOI to coordinated bimanual dexterous manipulation of up to two objects. We model multi-object interactions through dual denoising streams, coupled via bidirectional cross-attention and fused into a single coherent full-body motion. By modeling object motion through our hand-relative representation, we unify single-hand and bimanual manipulation within the same formulation, producing plausible hand-object interactions in single-shot inference, with no additional contact optimization. Our extensive evaluation shows that Dex2HOI generalizes across both single-object and multi-object settings, achieving \emph{SOTA} results and improved motion realism.

\section*{Acknowledgements}

S. Zafeiriou and part of the research was funded by the EPSRC Project
GNOMON (EP/X011364/1) and Turing AI Fellowship (EP/Z534699/1). RA Potamias was partially supported from Project GNOMON (EP/X011364/1).

\bibliography{references}
\bibliographystyle{references}

\clearpage
\pagebreak

\section*{Appendix}

In the \textbf{supplementary video} provided, we include a brief presentation of our method, contributions and evaluation, along with detailed \textbf{videos} of the figures shown in the paper. \textbf{Generated motions are best viewed in video form.} Training and evaluation code along with checkpoints will get released upon publication.

In this Appendix, we provide additional details to complement the main paper: Section~\ref{sec:impl_appendix} reports a full implementation breakdown, including architectural and training objective details. Section~\ref{sec:furthereval_appendix} presents \emph{further evaluation}, with extended results on GRAB ~\citep{taheri2020grab} under the LatentHOI ~\citep{Muchen_LatentHOI} physics protocol and additional quantitative and qualitative comparisons on HIMO ~\citep{lv2024himonewbenchmarkfullbody} and HUMOTO ~\citep{humoto_dataset} datasets. We also show examples of unseen object geometry adaptations and hand dexterity evaluation along with examples of multi- object HOI generation. Section~\ref{sec:perc_appendix} details our \emph{perceptual study analysis}, VLM and User based. We close with a {discussion of limitations and future directions} in Section~\ref{sec:disc_appendix}.

\section{Motion Generation with Diffusion Models - Background}
\label{sec:diffusion_background}

\paragraph{Diffusion-based Motion Generation.}
Denoising diffusion models formulate motion generation as the iterative reversal of a Gaussian corruption process. Given a clean motion sequence
$x_0 \in \mathbb{R}^{T \times D}$, the forward diffusion process progressively perturbs the motion with Gaussian noise according to
\begin{equation}
q(x_\tau \mid x_0)
=
\mathcal{N}
\left(
x_\tau;
\sqrt{\bar{\alpha}_\tau}x_0,
(1-\bar{\alpha}_\tau)I
\right),
\end{equation}
or equivalently,
\begin{equation}
x_\tau
=
\sqrt{\bar{\alpha}_\tau}\,x_0
+
\sqrt{1-\bar{\alpha}_\tau}\,\epsilon,
\qquad
\epsilon \sim \mathcal{N}(0,I),
\label{eq:forward_diffusion}
\end{equation}
where $\tau \in \{1,\ldots,N\}$ denotes the diffusion timestep,
$\alpha_\tau = 1-\beta_\tau$, and
$\bar{\alpha}_\tau = \prod_{s=1}^{\tau}\alpha_s$.
As $\tau$ increases, the structured motion $x_0$ is gradually transformed toward Gaussian noise.

The reverse process learns a conditional denoiser $D_\theta$ that reconstructs the clean motion,
\begin{equation}
\hat{x}_0
=
D_\theta(x_\tau,c,\tau),
\label{eq:reverse_diffusion}
\end{equation}
where $c$ denotes the generation conditions - in our case text and object geometry. Following common motion diffusion formulations, the model is trained by sampling a diffusion timestep $\tau$ and noise $\epsilon$, constructing $x_\tau$ according to Eq.~\ref{eq:forward_diffusion}, and minimizing the reconstruction objective
\begin{equation}
\mathcal{L}_{\mathrm{diff}}
=
\mathbb{E}_{x_0,\tau,\epsilon}
\left[
\left\|
x_0-D_\theta(x_\tau,c,\tau)
\right\|_2^2
\right].
\label{eq:diffusion_loss}
\end{equation}

Conceptually, the diffusion process can be summarized as:
\begin{equation}
\boxed{x_0}
\;\xrightarrow[\text{forward diffusion}]{+\epsilon}\;
\boxed{x_\tau}
\;\xrightarrow[\text{text + BPS}]{D_\theta}\;
\boxed{\hat{x}_0}.
\label{eq:diffusion_overview}
\end{equation}

In \textbf{Dex2HOI}, this formulation is extended to two coordinated object-specific diffusion streams. Each stream denoises the interaction between the human and one object, while bidirectional cross-attention layers enable information exchange between the streams:
\begin{equation}
\boxed{x_\tau^{A}}
\;\xleftrightarrow[\text{cross-attention}]{}
\boxed{x_\tau^{B}}
\;\xrightarrow{D_\theta}\;
\left\{
\begin{array}{l}
[\hat{h}^{A},\hat{r}^{A},\hat{o}^{A}]\\[2pt]
[\hat{h}^{B},\hat{r}^{B},\hat{o}^{B}]
\end{array}
\right\}.
\;\xrightarrow{D_h}\;
\boxed{
[\hat{h},\hat{r},\hat{o}^{A},\hat{o}^{B}]
}.
\label{eq:dex2hoi_diffusion_overview}
\end{equation}

The two streams independently model object-specific interactions while sharing information throughout denoising. The Motion Fusion Network $D_h$ unifies the two predicted human representations into a single body motion, while retaining the object motion predicted by each corresponding stream (only the body pose gets fused). To further support our unified one and two object training regime, when only one object is present the remaining stream and Motion Fusion Network $D_h$ are masked and not used for prediction. 

\section{Implementation Details}\label{sec:impl_appendix}

\paragraph{Implementation details.} Our two-object model uses a shared Transformer decoder with 8 layers, together with 8 cross-attention layers. All decoder and cross-attention blocks use a latent dimension of 512, 4 attention heads, and a feed-forward dimension of 1024. Object conditioning is provided by a shared BPS encoder ~\citep{prokudin2019efficient}. For the Motion Fusion Network, the two body streams $h_A, h_B$ are represented as a 2-token sequence at each frame and fused with a lightweight self-attention head using 4 attention heads; For HOI diffusion $D_{\theta}$, the model is trained with 10 diffusion steps, a cosine noise schedule, and fixed-small variance, and inference uses DDIM-10 sampling. Training uses EMA and AdamW with a learning rate of 1e-4 and 50-100k training steps, depending on the dataset. All training and evaluation experiments were executed between two NVIDIA GeForce RTX 4090 GPUs.

\paragraph{Classifier-Free Guidance.}
We apply classifier-free guidance (CFG) ~\citep{ho2022classifierfreediffusionguidance}  only to the text input conditions. During training, the encoded text tokens are randomly zeroed with probability \(p_{\mathrm{drop}}\), while all geometric and temporal inputs remain unchanged. At sampling time, we run the denoiser twice: once with the full text prompt and once with the text embedding masked out. The two predictions are combined using the guidance scale \(w\), which amplifies the direction induced by the prompt. Importantly, the unconditional branch keeps the same object point-cloud embeddings, object masks, prefix frames, initial transforms, and optional root/object \(x_0\) position conditions as the conditional branch. Thus CFG improves semantic prompt alignment while preserving the scene geometry and first-frame constraints, without adding extra networks or training objectives.

\paragraph{Training Objective Analysis.}

In Section \ref{sec:trainobjective}, we define our full training objective as:
\begin{equation}
\mathcal{L}
=
\mathcal{L}_{diff}
+ \mathcal{L}_{pos} + \mathcal{L}_{vel} + 
\underbrace{
\lambda_{w}\mathcal{L}_{mix}
+
\lambda_{t}\mathcal{L}_{trans}
+
\lambda_{o}\mathcal{L}_{off}
}_{\text{\emph{object representation supervision}}} +
\lambda_{p}\mathcal{L}_{dist}
+
\lambda_{pen}\mathcal{L}_{pen}
\end{equation}

\noindent\textbf{Velocity and Position Supervision.}
Beyond the denoising objective $\mathcal{L}_{diff}$ we add first-order velocity
matching losses to suppress temporal jitter and enforce smooth dynamics
across the generated motion.
We compute the following geometric losses through forward kinematics,
which converts the predicted body poses (6D rotations) to 3D joint positions.

\smallskip
\noindent\textit{Joint Position Loss.}
The global joint position loss $\mathcal{L}_{pos}$ penalises discrepancies
in the predicted 3D locations of body joints:
\begin{equation}
\mathcal{L}_{pos}(\mathbf{p},\hat{\mathbf{p}})
= \|\mathbf{p} - \hat{\mathbf{p}}\|_{2}^{2},
\end{equation}
where $\mathbf{p}\in\mathbb{R}^{T\times J\times 3}$ are the ground-truth
global joint positions with $J$ joints, and $\hat{\mathbf{p}}$ are the
corresponding predicted positions.
This loss enforces spatial accuracy in the generated motions.

\smallskip
\noindent\textit{Joint Velocity Loss.}
To promote temporal smoothness and physical plausibility, we apply a velocity
loss between the ground truth $\mathbf{v}$ and predicted $\hat{\mathbf{v}}$ joint positions:
\begin{equation}
\mathcal{L}_{vel}^{J}(\mathbf{v},\hat{\mathbf{v}})
= \|\mathbf{v} - \hat{\mathbf{v}}\|_{2}^{2},
\end{equation}
where $\mathbf{v}_{t} = \mathbf{p}_{t} - \mathbf{p}_{t-1}$ represents the
joint velocities computed as the difference between consecutive frames.
This loss discourages unnatural jittering and encourages smooth,
realistic motion trajectories.

\smallskip
\noindent\textit{Hand-Object Penetration Loss.}
To penalize fingertip interpenetration with the object surface, we apply a
penalty on the signed distance field evaluated at predicted fingertip positions
$\hat{\mathbf{h}}$:
\begin{equation}
\mathcal{L}_{pen}^{HO}(\hat{\mathbf{h}})
= \max\bigl(0,\; -\mathrm{SDF}(\hat{\mathbf{h}})\bigr),
\end{equation}
where $\mathrm{SDF}(\cdot)$ returns positive values in free space and negative
values inside the object. This loss fires only when a fingertip penetrates the
object surface, pushing predicted joints back to the exterior.

\smallskip
\noindent\textit{Body-Object Penetration Loss.}
An analogous penalty is applied to the full body mesh vertices
$\mathbf{v}$ to prevent torso and limb collisions with the object:
\begin{equation}
\mathcal{L}_{pen}^{BO}(\mathbf{v})
= \max\bigl(0,\; - \mathrm{SDF}(\mathbf{v})\bigr).
\end{equation}

\smallskip

\noindent\textit{Object-Object Penetration Loss.}
To address the possibility of penetration between objects - when simultaneous manipulation is involved - the same criterion is applied
between the objects themselves. Let $\mathcal{P}_{k}$ denote a set of points sampled on the surface of object $k$, expressed in world coordinates under its predicted pose. Then the penetration penalty is symmetric in the two objects:
\begin{equation}
\mathcal{L}_{pen}^{OO}
= \frac{1}{|\mathcal{P}_{b}|}\sum_{\mathbf{p}\in\mathcal{P}_{b}}
  \max\bigl(0,\,-\mathrm{SDF}_{a}(\mathbf{p})\bigr)
\;+\;
\frac{1}{|\mathcal{P}_{a}|}\sum_{\mathbf{p}\in\mathcal{P}_{a}}
  \max\bigl(0,\,-\mathrm{SDF}_{b}(\mathbf{p})\bigr),
\end{equation}
where $\mathrm{SDF}_{k}(\cdot)$ is the signed distance field of object $k$.

Empirically, in our experiments we set the object-representation supervision
lambdas as $\lambda_{w} = \lambda_{o} = 0.5$, the object translation weight
$\lambda_{t} = 1.0$, while for the position and velocity losses we use $\lambda_{pos} = 1.0$ and $\lambda_{vel} = 1.0$ respectively. Finally, we use a penetration weight of $\lambda_{pen} = 1.0$ across penetration terms, and $\lambda_{p} = 2.0$ for our hand-object contact regression loss.

\begin{figure}[h]
  \centering
  \includegraphics[width=1\linewidth]{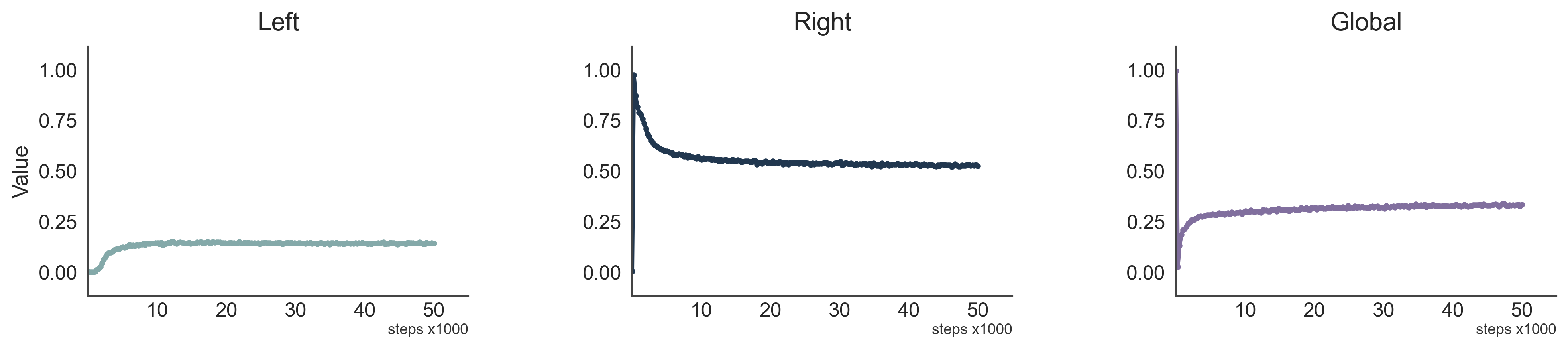}
  \includegraphics[width=1\linewidth]{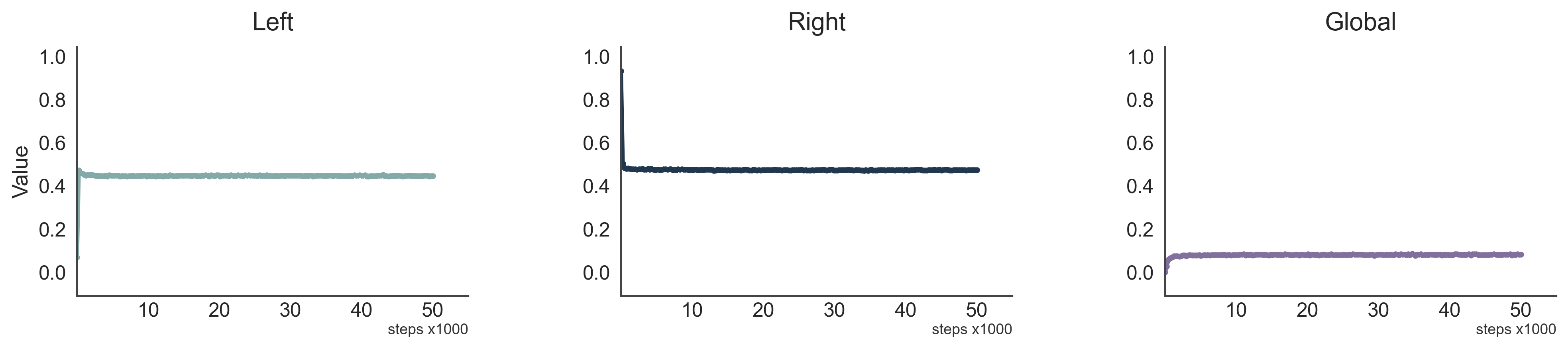}
  \caption{Average predicted mixture weights ($w$) over training for the Hand-Relative Object Representation. Left/right denote the hand-conditioned object branches, and global denotes the global branch for GRAB (top) and HUMOTO (bottom) datasets.}
  \label{fig:weights}
\end{figure}

\smallskip
\noindent\textit{Ground-truth mixture weights.}
The ground-truth targets $[w^L_t, w^R_t, w^G_t]$ are derived from hand
proximity to the object at each frame.
For each hand $s \in \{L, R\}$ we measure the Euclidean distance $d^s_t$
from the wrist center to the object center, 
\begin{wrapfigure}[16]{r}{0.4\linewidth}
  \centering
  \includegraphics[width=\linewidth]{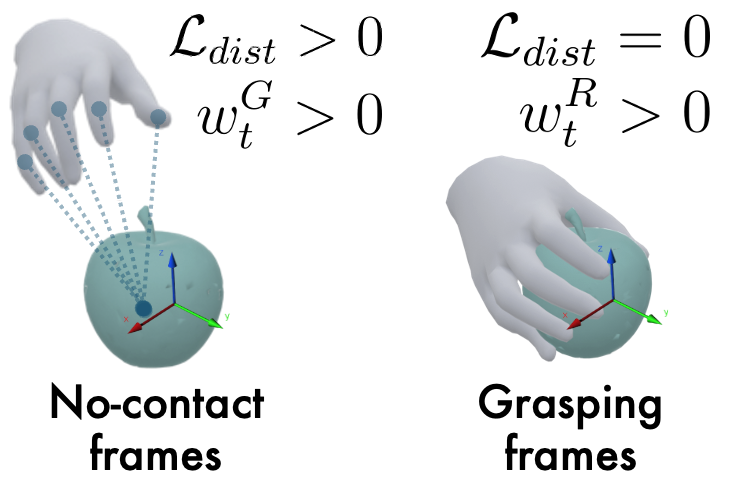}
  \caption{Object Motion Paradigm; No-contact frames have positive $\mathcal{L}_{Dist}$ and use $w_t^G$. Contact frames have $w_t^G\approx0$.}
  \label{fig:sample}
\end{wrapfigure}
then compute gated contact score:
\begin{equation}
  \phi^s_t = \sigma\!\left(r_c - d^s_t\right),
  \qquad s \in \{L, R\},
\end{equation}

where $r_c = 0.15\,\mathrm{m}$ is the contact radius ($\phi^s_t \approx 1$ when the hand is within $r_c$,
$\phi^s_t \approx 0$ otherwise).
The global score captures frames where neither hand controls the object:
$\phi^G_t = (1 - \phi^L_t)(1 - \phi^R_t)$.
The three scores are $\ell^1$-normalised to obtain the final weights:
\begin{equation}
  \bigl[w^L_t,\, w^R_t,\, w^G_t\bigr]
  =
  \frac{1}{\phi^L_t + \phi^R_t + \phi^G_t}
  \bigl[\phi^L_t,\, \phi^R_t,\, \phi^G_t\bigr].
\end{equation}
When only the left hand is near the object $w^L \approx 1$;
when both hands are near $w^L \approx w^R \approx 0.5$; and when neither
hand is close $w^G \approx 1$, delegating object position to the free
global branch. Figure~\ref{fig:weights} shows the \textbf{mean} weight regression across training on GRAB (top) and HUMOTO (bottom).

\section{Further Evaluation}\label{sec:furthereval_appendix}

\textbf{Metrics Analysis.} As in the main paper, we evaluate generated motions along three axes: distributional realism, text–motion alignment, and physical plausibility. Namely, FID measures the distance between the distribution of generated and real motion embeddings. Diversity measures the spread of generated motions in embedding space. MultiModality (MM) quantifies how varied the model's outputs are for the same text prompt. R-Precision@3 measures top-3 motion-to-text retrieval accuracy from a pool of candidates (we use size 32). MMdist measures the alignment between each generated motion and its conditioning text in the shared embedding space. Finally, penetration depth quantifies physical plausibility by measuring how object vertices intersect the body mesh. 

\subsection{More Evaluation on the GRAB Dataset} 

To further assess the quality of the contacts produced by \textbf{Dex2HOI}, we compare against hand-centric grasping method LatentHOI~\citep{Muchen_LatentHOI}, which focuses exclusively on dexterous hand–object interaction without modeling the full body. This method is not evaluated on motion representation metrics (FID, R-Precision, etc.), and instead uses a physics-inspired protocol, explained below.

To enable a fair comparison and also test our method's ability to generalize beyond known objects, we retrain Dex2HOI on the GRAB dataset following the unseen-object protocol of LatentHOI: the four held-out object categories (\textit{apple}, \textit{mug}, \textit{train}, \textit{elephant}) are excluded from training and used exclusively for evaluation. This setting probes whether our hand-relative representation produces grasps that transfer to novel geometries, rather than overfitting to learned object instances.

\begin{figure}[h]
  \centering
  \includegraphics[width=1\linewidth]{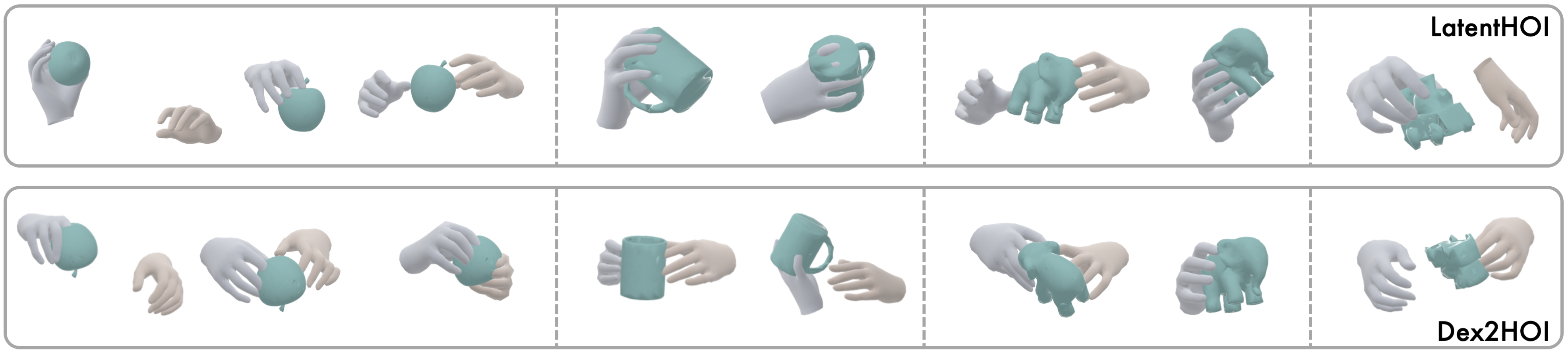}
  \caption{\textbf{Dex2HOI} vs LatentHOI~\citep{Muchen_LatentHOI} grasping generations on 4 unseen object geometries.}
  \label{fig:latenthoi}
\end{figure}

\begin{figure}[p]
  \centering
  \includegraphics[width=1\linewidth]{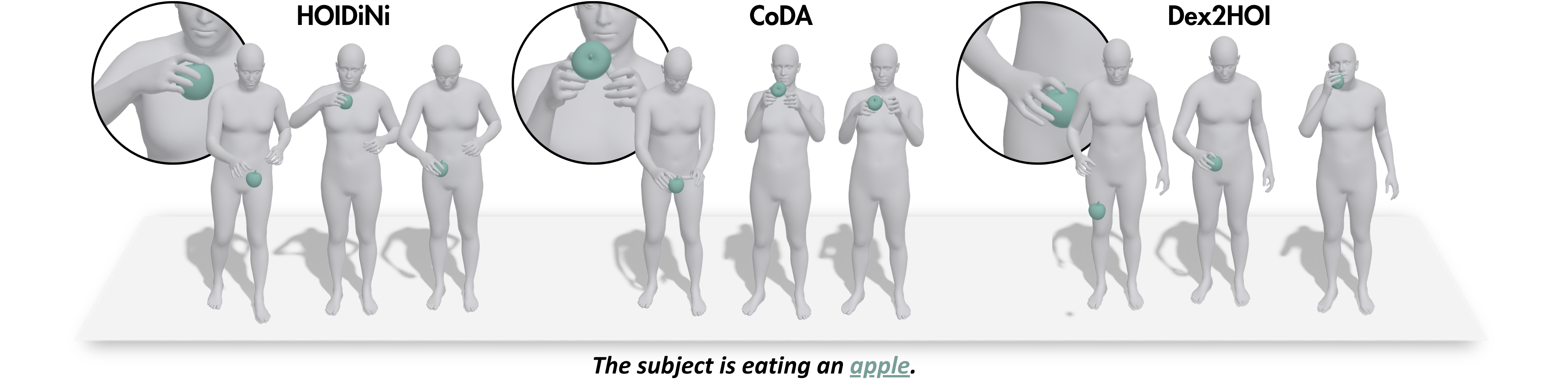}
  \includegraphics[width=1\linewidth]{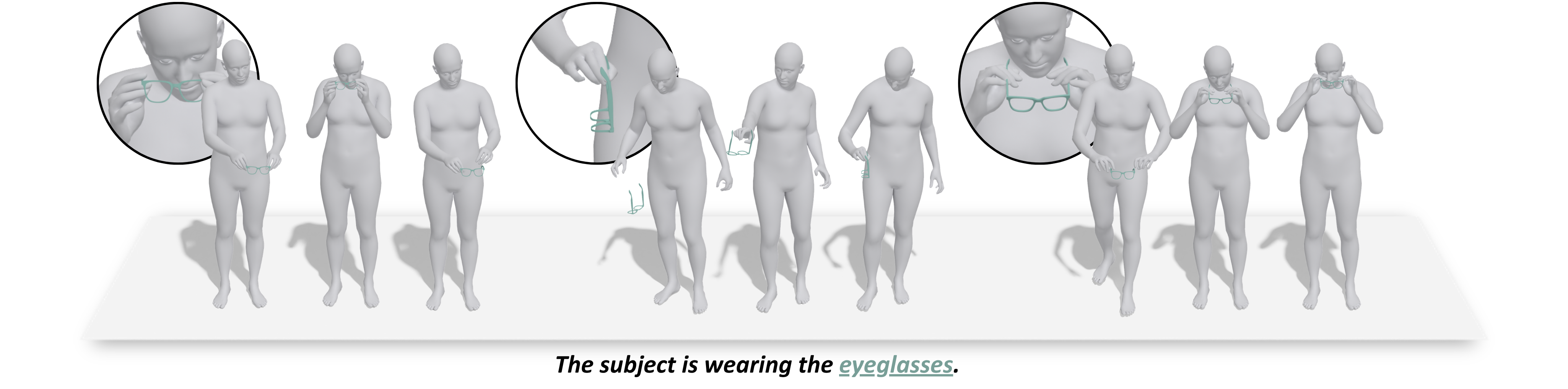}
  \includegraphics[width=1\linewidth]{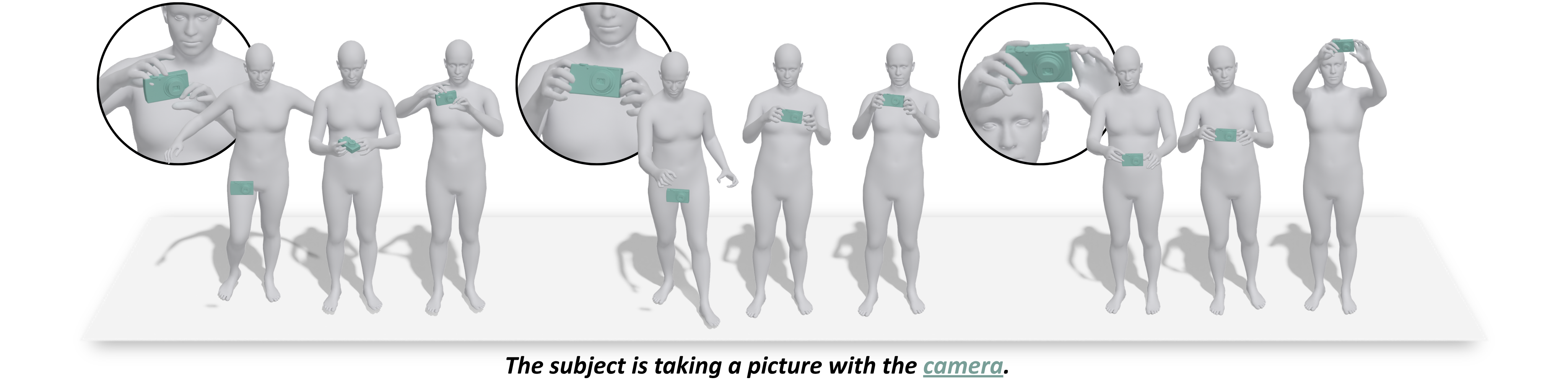}
    \includegraphics[width=1\linewidth]{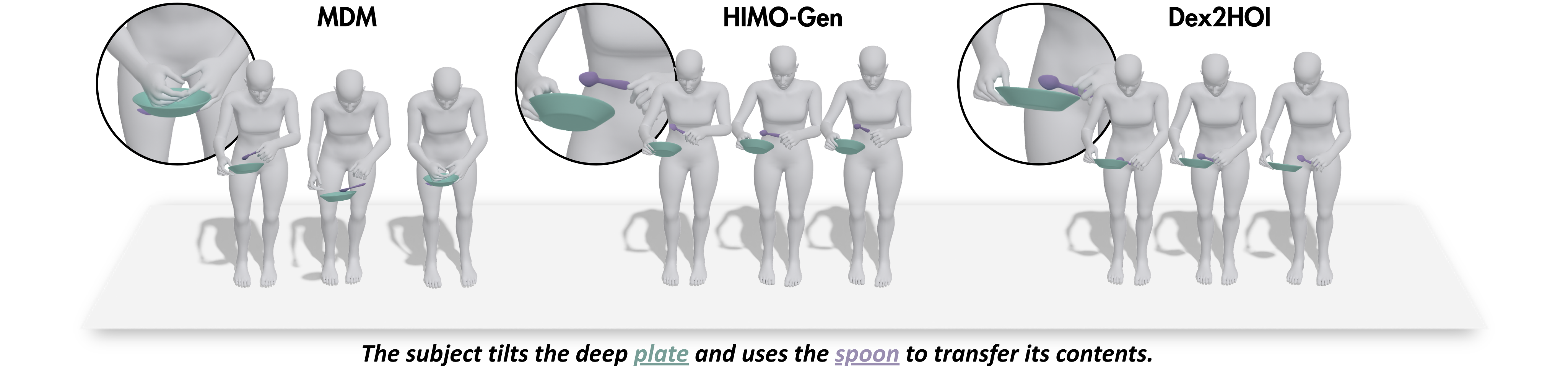}
  \includegraphics[width=1\linewidth]{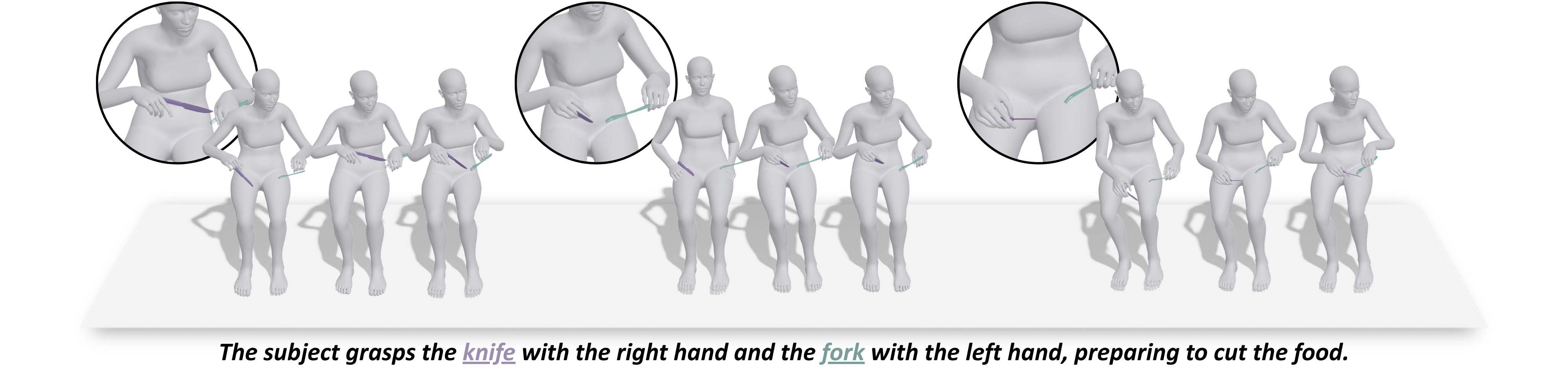}
  \includegraphics[width=1\linewidth]{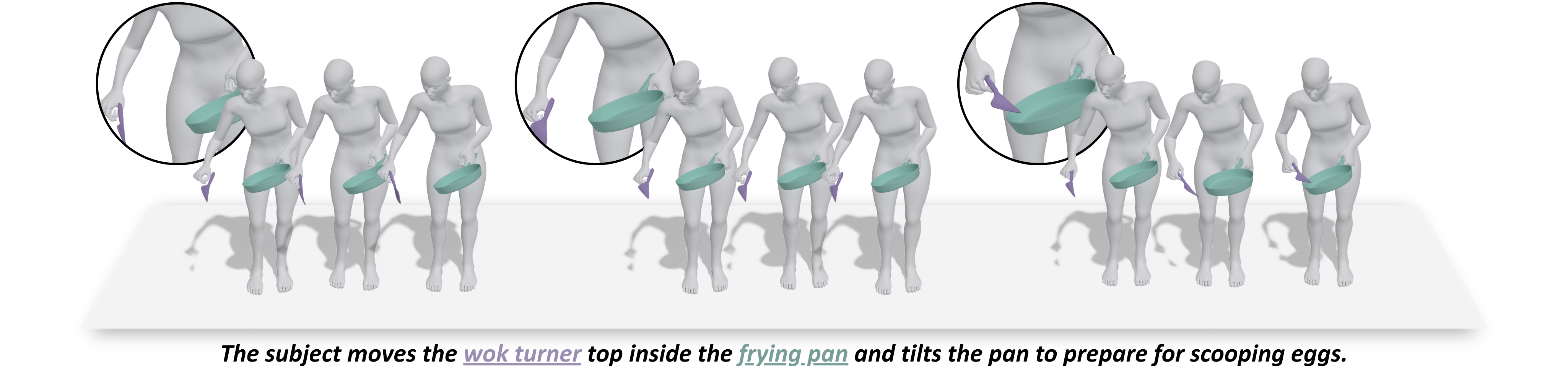}
  \caption{Additional qualitative comparisons of \textbf{Dex2HOI (right)} on the GRAB dataset, against HOIDiNi (left) and CoDA (middle), and on HUMOTO (right), against HOIDiNi (left) and CoDA (middle). Dex2HOI generates higher-quality interactions in single-shot.}
  \label{fig:grab_fig_supplementary}
\end{figure}

We adopt the LatentHOI~\citep{Muchen_LatentHOI} physics protocol: interpenetration volume (IV, $cm^3$) and depth (ID, $cm$) measure hand–object mesh overlap; contact ratio (CR) measures the fraction of hand vertices within contact distance of the object; IVU is the normalized in-volume fraction over off-ground frames; Phy is the percentage of off-ground frames with positive hand–object contact; and SD and OD measure per-prompt and overall motion diversity. For the hand-only comparison, we report mesh-based hand–object penetration computed bidirectionally between the hand mesh and the object mesh, so that both hand-inside-object and object-inside-hand failure modes are captured. Diversity is computed across subjects performing the same action (inter-subject variation). Table~\ref{tab:grab_eval_latenthoi} reports the results, with \textbf{bold} and \underline{underlined} entries indicating the best and second-best generative models respectively. Finally, Figure ~\ref{fig:latenthoi} compares generated grasps on the 4 unseen object geometries, while Figure ~\ref{fig:grab_fig_supplementary} provides additional qualitative results on GRAB, against CoDA and HOIDiNi. 

\begin{table*}[h]
\centering
\resizebox{\linewidth}{!}{%
\begin{tabular}{ll c cc cc cc c cc}

& & \multicolumn{2}{c}{IV [cm$^3$] $\downarrow$} & \multicolumn{2}{c}{ID [cm] $\downarrow$} & \multicolumn{2}{c}{CR [\%] $\uparrow$} &  &   &    & \\
\cmidrule(lr){3-4}\cmidrule(lr){5-6}\cmidrule(lr){7-8}
Method & & Right & Left & Right & Left & Right & Left & IVU $\downarrow$ & Phy $\uparrow$ & SD $\uparrow$ & OD $\rightarrow$ \\
\midrule
GT  & & 4.97  & 1.20 & 0.92 & 0.18 & 8.21  & 1.56  & 0.12 & 98.68 & --   & 0.40 \\
\cdashline{1-12}\noalign{\vskip 2pt}
IMoS~~\citep{ghosh2022imos}  & & 10.38 & --   & 1.25 & --   & 4.61  & --    & 0.53 & {84.88} & 0.00 & \textbf{0.46} \\
MDM~~\citep{mdm_tevet2023human}       & & 9.12  & 2.61 & 1.24 & 0.51 & 8.21  & \underline{1.29}  & \underline{0.19} & \underline{89.81} & \underline{0.18} & 0.29 \\
MLD~~\citep{chen2023executing_mld}    & & 9.62  & 3.14 & \underline{1.06} & \underline{0.49} & 10.23 & 0.87  & 0.24 & 85.68 & {0.20} & 0.31 \\
LatentHOI~~\citep{Muchen_LatentHOI} & & \textbf{6.38} & \textbf{1.66} & \textbf{0.77} & \textbf{0.29} & \underline{11.94} & {1.11} & \textbf{0.10} & \textbf{96.16} & 0.13 & 0.27 \\
\rowcolor{trowcolor}
	\textbf{Dex2HOI} & & \underline{7.39} & \underline{2.59} & 1.94 & 0.75 & \textbf{15.02} & \textbf{7.35} & 0.26 & 81.98 & \textbf{0.20} & \underline{0.47} \\
\bottomrule
\end{tabular}}
\smallskip
\raggedright
\caption{%
  	{GRAB hand–object interaction evaluation against LatentHOI~\citep{Muchen_LatentHOI}.} 
}
\label{tab:grab_eval_latenthoi}
\end{table*}

\clearpage
\pagebreak

\clearpage
\pagebreak

\subsection{HUMOTO Unseen Geometries and Dexterity study}
\label{sec:dexdex}

\paragraph{HUMOTO Two-Object Test Split.}
Throughout our experiments we use a split that contains 45 two-object sequences spanning seven scene domains (since the original HUMOTO dataset does not come with a pre-made split). Kitchen and cooking interactions constitute the largest subset (27/45, 60\%), followed by office (6/45, 13.3\%), dining-room (4/45, 8.9\%), living-room (3/45, 6.7\%), bedroom (3/45, 6.7\%), garage (1/45, 2.2\%), and outdoor (1/45, 2.2\%) scenarios. The interactions are manipulation-centric: grasping is the most frequent dominant action, followed by placement and transfer operations including moving, placing, putting, transferring, and pouring. Across the 90 moving-object instances (45 sequences $\times$ two objects), the most frequent objects are mug (12), spoon (9), wok turner (9), deep plate (8), and frying pan (7), while 15 additional object categories occur only once. The full train/test split will be released upon publication to facilitate future research on motion-generation tasks involving diverse HUMOTO objects and actions.

\begin{figure}[h]
  \centering
  \includegraphics[width=1\linewidth]{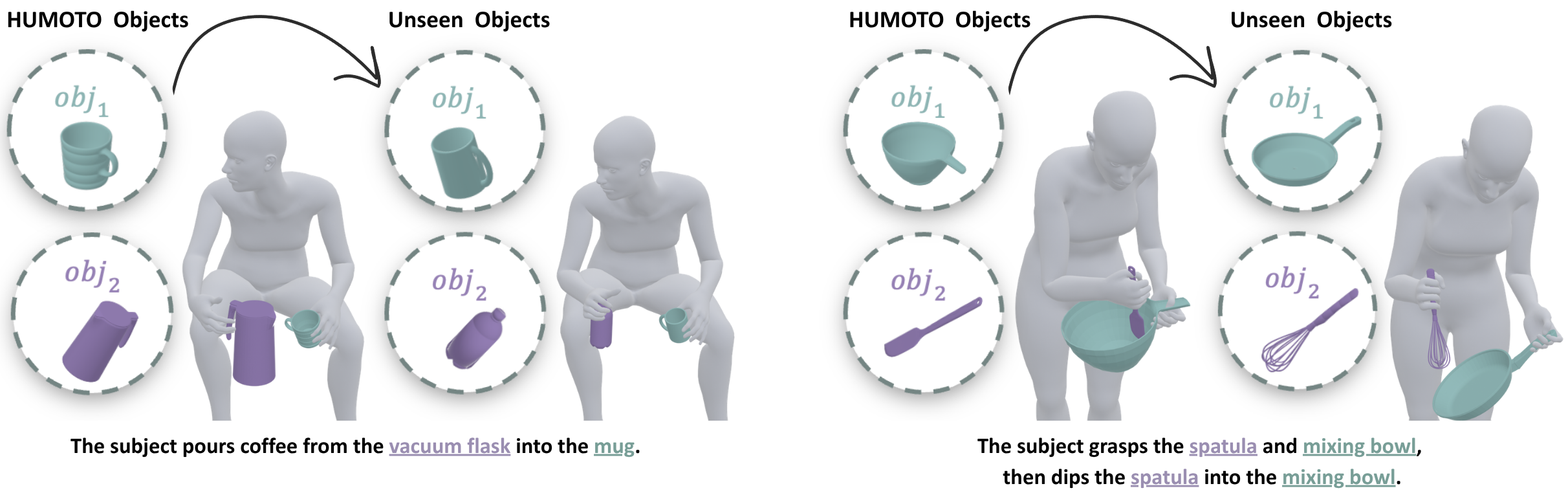}
 \caption{\textbf{Zero-shot object switching on HUMOTO.}
}
  \label{fig:unseen_objects}
\end{figure}

To assess whether Dex2HOI learns transferable interaction priors rather than memorizing the object geometries observed during training, we perform an additional zero-shot object-switching experiment. Specifically, we replace the original HUMOTO object geometries at inference with semantically related but geometrically unseen objects sourced from HIMO, while preserving the interaction prompt and generating the motion without retraining or test-time optimization. As illustrated in Fig.~\ref{fig:unseen_objects}, Dex2HOI adapts the generated interaction to substantial geometric changes, including replacing a vacuum flask and mug with unseen container geometries, and replacing a mixing bowl and spatula with a frying pan and whisk.

\begin{table*}[h]
\centering
\small
\setlength{\tabcolsep}{8pt}

\label{tab:finger_diversity_humoto}
\begin{tabular}{lcccc}
\toprule
\textbf{Diversity} & \textbf{GT} & \textbf{MDM} & \textbf{HIMO-Gen} & \textbf{Dex2HOI} \\
\midrule
Left Hand World
& 16.037 & 15.120 $\pm$ 0.082 & 14.228 $\pm$ 0.103 & \textbf{15.552 $\pm$ 0.041} \\
Left Hand Wrist-relative
& 3.371 & 3.125 $\pm$ 0.020 & 2.951 $\pm$ 0.024 & \textbf{3.368 $\pm$ 0.012} \\
\quad Left Thumb
& 1.118 & 1.054 $\pm$ 0.006 & 1.011 $\pm$ 0.009 & \textbf{1.115 $\pm$ 0.004} \\
\quad Left Index
& 1.667 & 1.567 $\pm$ 0.008 & 1.491 $\pm$ 0.014 & \textbf{1.669 $\pm$ 0.005} \\
\quad Left Middle
& 1.677 & 1.568 $\pm$ 0.009 & 1.466 $\pm$ 0.013 & \textbf{1.673 $\pm$ 0.005} \\
\quad Left Ring
& 1.534 & 1.424 $\pm$ 0.011 & 1.320 $\pm$ 0.012 & \textbf{1.531 $\pm$ 0.006} \\
\quad Left Pinky
& 1.424 & 1.316 $\pm$ 0.012 & 1.206 $\pm$ 0.010 & \textbf{1.421 $\pm$ 0.005} \\
\midrule
Right Hand World
& 16.965 & 15.851 $\pm$ 0.076 & 14.975 $\pm$ 0.104 & \textbf{16.762 $\pm$ 0.038} \\
Right Hand Wrist-relative
& 3.287 & 3.067 $\pm$ 0.012 & 2.876 $\pm$ 0.029 & \textbf{3.284 $\pm$ 0.010} \\
\quad Right Thumb
& 1.235 & 1.124 $\pm$ 0.006 & 1.044 $\pm$ 0.014 & \textbf{1.238 $\pm$ 0.004} \\
\quad Right Index
& 1.711 & 1.590 $\pm$ 0.006 & 1.505 $\pm$ 0.017 & \textbf{1.708 $\pm$ 0.005} \\
\quad Right Middle
& 1.571 & 1.486 $\pm$ 0.006 & 1.407 $\pm$ 0.014 & \textbf{1.573 $\pm$ 0.004} \\
\quad Right Ring
& 1.419 & 1.320 $\pm$ 0.005 & 1.247 $\pm$ 0.011 & \textbf{1.416 $\pm$ 0.004} \\
\quad Right Pinky
& 1.306 & 1.211 $\pm$ 0.004 & 1.129 $\pm$ 0.011 & \textbf{1.308 $\pm$ 0.003} \\
\bottomrule
\end{tabular}
\caption{\textbf{Hand-motion diversity on the HUMOTO test set.}
World-space hand diversity captures both global hand motion and articulation, whereas wrist-relative diversity removes global wrist displacement and therefore more directly measures local hand articulation. Per-finger results further isolate individual finger motion. Dex2HOI consistently remains closest to the ground-truth diversity.}
\end{table*}

\subsection{More Evaluation on HIMO Dataset} 

\paragraph{HIMO.} For our next experiments, we use the 2-object split ~\citep{lv2024himonewbenchmarkfullbody}, which contains 2,578 training and 522 test sequences, to assess dexterous bimanual two-object motion generation.  

\paragraph{HIMO Motion Evaluator.} Following our previous protocol, we train a dedicated HIMO motion evaluator with a contrastive text–motion objective on the 2-object split, enabling the computation of FID, R-Precision, MMdist, Diversity, and Multimodality maintaining consistency with prior text-to-motion benchmarks. This evaluator is used to compute all distributional metrics reported in Table~\ref{tab:himoeval_sup}, where Dex2HOI again surpasses baselines HIMO-Gen and MDM.

\paragraph{HIMO vs. HUMOTO Ground-Truth Quality.} While HIMO provides a valuable benchmark for two-object manipulation, ground-truth motions exhibit noticeable contact-displacement artifacts — objects occasionally drift away from the hand during manipulation (as seen in the GT Motion section of Figure~\ref{fig:himosup}). For this reason, we refrain from showing generated videos on HIMO in the supplementary video, as these artifacts in the training data propagate to the generated outputs and are difficult to interpret in isolation. Quantitatively, however, Dex2HOI still produces faithful hand-object following: as reported in Table~\ref{tab:himoeval_sup}, our method achieves the best FID (0.46), R-Precision@3 (0.369), and penetration (9.15mm), indicating that the model captures the underlying interaction structure despite the noisier supervision signal. Furthermore, as shown in the qualitative generated examples provided in Figure~\ref{fig:himosup}, Dex2HOI produces coordinated hand-object contacts that align well with the instruction prompt.

\begin{table}[!htbp]
\centering

\resizebox{\linewidth}{!}{%
\begin{tabular}{l|lcccccc}
\toprule
 & Method & FID $\downarrow$ & Diversity $\rightarrow$ & MModality $\uparrow$ & Pen.\ (mm) $\downarrow$ & R-Pre@3 $\uparrow$ & MMdist $\downarrow$ \\
\midrule
& GT   & --- & 1.184 & --- & --- & 0.791 & 1.029 \\
\cdashline{2-8}\noalign{\vskip 2pt}
& MDM ~\citep{mdm_tevet2023human}  & \underline{0.54} & \textbf{1.097} & \textbf{0.628} & 13.22 & \underline{0.295} & \textbf{1.083} \\
& HIMO-Gen ~\citep{lv2024himonewbenchmarkfullbody} & 1.09 & 0.445 & 0.323 & \underline{9.95} & 0.140 & 1.393 \\
\rowcolor{trowcolor}
\multirow{-4}{*}{\rotatebox{90}{2 objects}}  \cellcolor{white} &  \textbf{Dex2HOI}  & \textbf{0.46} & \underline{0.865} & \underline{0.470} & \textbf{9.15} & \textbf{0.369} & \underline{1.280} \\
\hline
\end{tabular}}
\caption{Evaluation of \textbf{HIMO} \emph{two-object} dataset. We highlight \textbf{best} and \underline{second best} results.}
\label{tab:himoeval_sup}
\end{table}

\begin{figure}[t]
  \centering
  \includegraphics[width=1\linewidth]{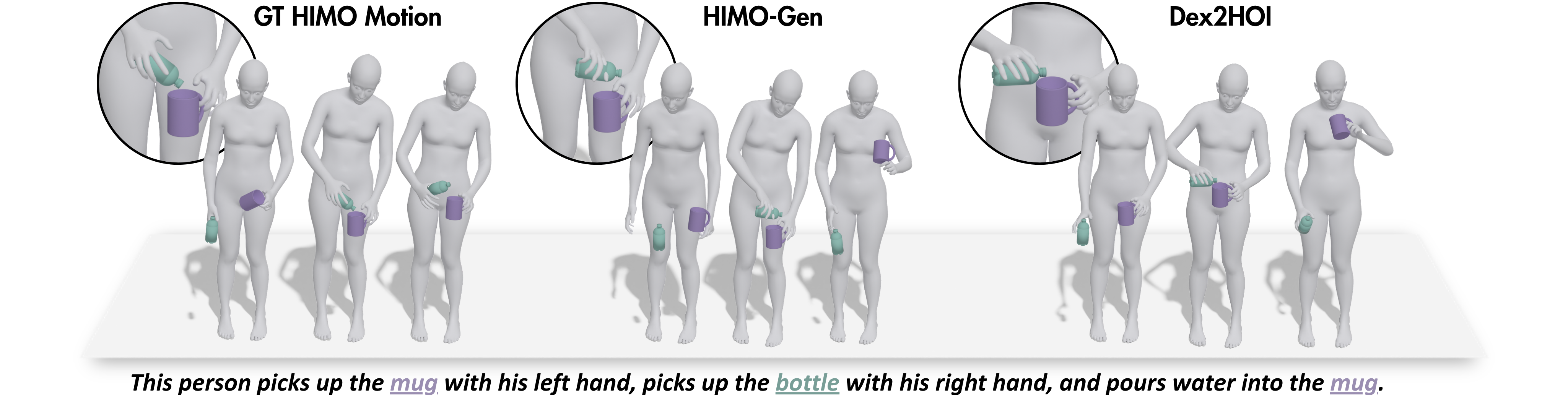}
  \includegraphics[width=1\linewidth]{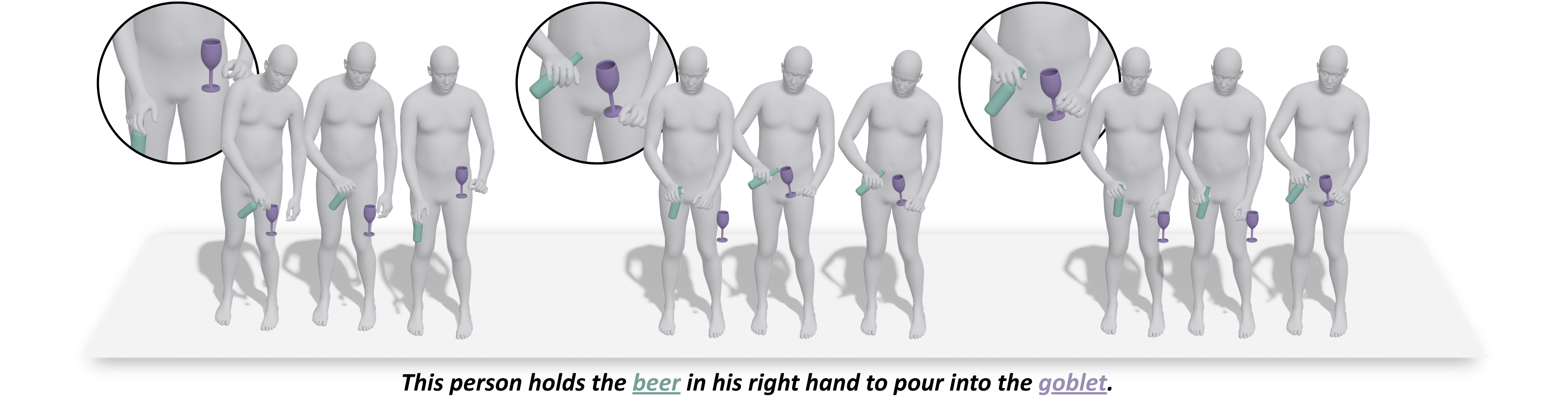}
  \caption{Additional qualitative results on the HIMO dataset~\citep{lv2024himonewbenchmarkfullbody}; We show Dex2HOI generations (right) against HIMO ground-truth motion (left).}
  \label{fig:himosup}
\end{figure}

\section{Expanding to More than Two Objects}
\label{sec:multi_object_extension}

Although Dex2HOI explicitly models up to two objects within each denoising window, its autoregressive formulation enables longer interactions involving a larger set of objects over time. We leverage this capability to experiment with longer horizon motion prediction and sequential interaction with more than two objects. In the case of HUMOTO (and HIMO) we can clearly distinguish between two types of multi-object interactions - both of which are included in the supplementary video provided: 

\paragraph{(1) sequential single-hand hand-offs,}where the subject is never in contact with all three objects at once - e.g. they set an object down before picking up the third - which is exactly a chained sequence of two-object interactions and can be represented as two consecutive two-object windows with a hand-off event, requiring no new architecture; and

\paragraph{(2) compositional carrying,}where multiple secondary objects are moved rigidly together with a base object (e.g. carrying a tray with glasses); here the secondary objects' motion can be determined by the base object's motion and can be captured by a single unified BPS representation rather than as independent streams.

\begin{figure}[h]
  \centering
  \includegraphics[width=\linewidth]{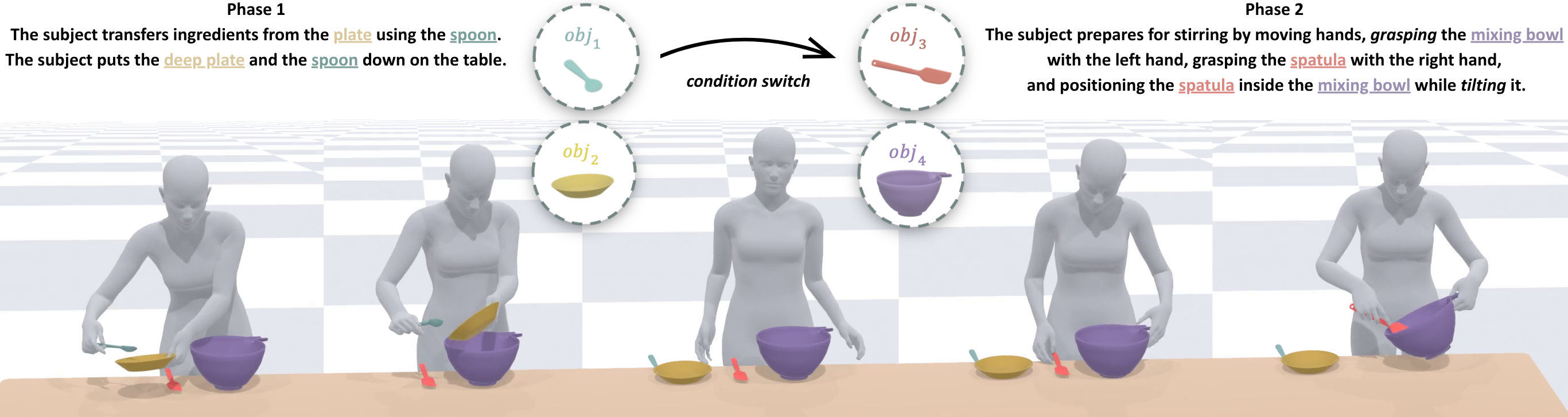}
  \caption{\textbf{Autoregressive generation beyond two objects.}
  While each Dex2HOI denoising window models at most two active objects, the object conditions can be switched between autoregressive windows. Previously generated motion is retained as the temporal prefix while new object geometries and interaction conditions are introduced, allowing a continuous sequence to compose interactions with more than two objects without modifying the model architecture or retraining.}
  \label{fig:multiobj}
\end{figure}

Figure~\ref{fig:multiobj} demonstrates this capability by switching the active object conditions throughout autoregressive generation, particularly in the HUMOTO.kitchen table setting.

\section{Perceptual Studies Analysis}\label{sec:perc_appendix}

To complement the quantitative metrics reported in Tab.~1, we conduct two
perceptual evaluation studies: an automated VLM-based evaluation and a
human preference study. Both are designed to probe dimensions of motion
quality that FID/diversity scores do not capture, namely perceptual
naturalness, semantic alignment with the text prompt, and hand-contact
plausibility.

\paragraph{Automated VLM Evaluation.}
We evaluate generated motions using a video-language model (VLM) to
obtain \textbf{scalable} assessment across all test sequences.  For
each generated motion we render a mesh-body video and present it to
\textbf{InternVL3-8B}~\citep{InternVL3} alongside three structured prompts:

\begin{enumerate}
  \item \textit{Naturalness}: "Rate how natural and human-like the body
    movement looks on a scale of 1 to 5, where 1 = completely
    unnatural/robotic and 5 = perfectly natural human motion."
  \item \textit{Semantic alignment}: "Rate how well the motion matches the
    description \texttt{\{prompt\}} on a scale of 1 to 5."
  \item \textit{Contact quality}: "Rate how well the hands make physical
    contact with the object on a scale of 1 to 5, where 1 = hands never
    touch the object and 5 = hands grasp the object convincingly."
\end{enumerate}

Each question is answered independently and the model is constrained to reply with a single integer from 1 to 5, making the evaluation deterministic. We run this protocol on two benchmarks;  On the \textbf{GRAB} test set we evaluate 145 test set sequences for each of CoDA, HOIDiNi, and Dex2HOI. On the \textbf{HUMOTO} test set we evaluate 45 sequences for MDM, HIMO-Gen, and Dex2HOI. To further validate our results, we run a \emph{pairwise} choice evaluation: for every pair of methods $(A, B)$, the model is shown both videos and chooses which exhibits better naturalness, contact, or overall quality.  

\begin{figure}[h]
  \centering

  \includegraphics[width=0.81\linewidth]{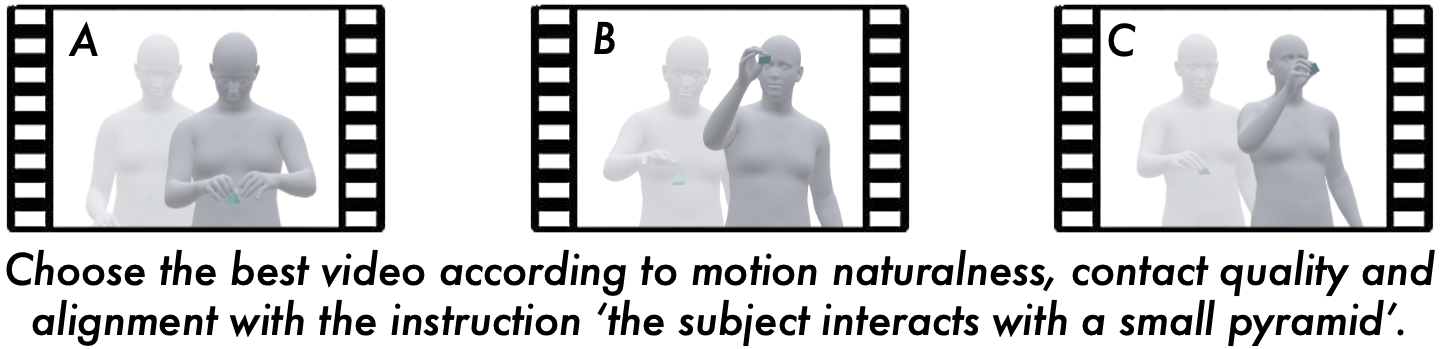}
    \caption{\textbf{User study}. A sample of the user study where 30 participants evaluated Dex2HOI generated motions against baseline methods.}
  \label{fig:usera}
\end{figure}

\begin{figure}[h]
  \centering
  \includegraphics[width=1\linewidth]{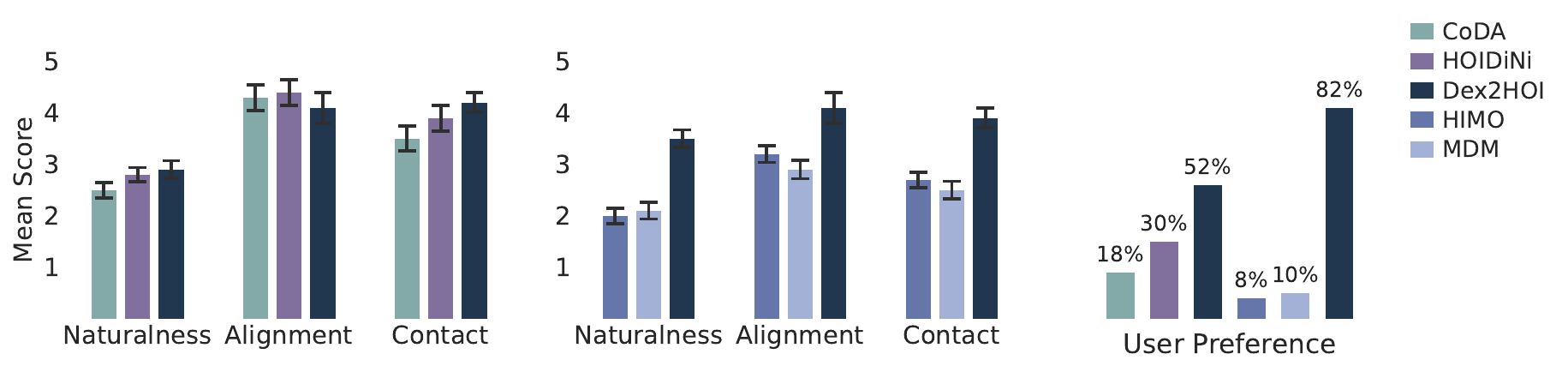}
  \caption{\textbf{Perceptual Studies}: \emph{Left}: VLM-eval for 1-object GRAB dataset. \emph{Middle}: VLM-eval for 2-object HUMOTO dataset. \emph{Right}: Collective User Preference. \textbf{Dex2HOI} achieves superior scores in realism and contact quality and is preferred by the majority of participants over competing approaches.}
  \label{fig:user}
\end{figure}

As shown in Figure~\ref{fig:user}, \textbf{Dex2HOI} consistently achieves superior interaction scores, in both single- and two-object benchmarks, demonstrating more natural motions and better contacts than competing baselines, further validated by our user study.

\section{Discussion and Broader Impact}\label{sec:disc_appendix}

\paragraph{Limitations.} Despite \textbf{Dex2HOI}'s promising results, certain challenges remain. While our architecture supports long sequence synthesis through autoregressive sampling, the diversity and complexity of the motions remain bounded by the bimanual interactions present in current 4D HOI datasets, which are limited. Future work should address long-horizon action planning and the development of standardized benchmarks for scaling multi-object bimanual manipulation.

\paragraph{Broader Impact.} Dex2HOI advances the synthesis of dexterous bimanual single- and two-object interactions, with potential to support a range of beneficial applications including immersive AR/VR experiences, animation pipelines, and the training of embodied agents and assistive robotic systems that operate in human-populated environments, while adhering to established ethical guidelines regarding synthetic data generation and use. 

\end{document}